\documentclass[sigconf, nonacm]{acmart}

\newcommand\vldbdoi{XX.XX/XXX.XX}
\newcommand\vldbpages{XXX-XXX}
\newcommand\vldbvolume{17}
\newcommand\vldbissue{7}
\newcommand\vldbyear{2024}
\newcommand\vldbauthors{\authors}
\newcommand\vldbtitle{\shorttitle} 
\newcommand\vldbavailabilityurl{https://github.com/zshhans/MSD-Mixer}
\newcommand\vldbpagestyle{empty}

\usepackage{algorithmic}
\usepackage{graphicx}
\usepackage{textcomp}
\usepackage{xcolor}
\usepackage{hyperref}
\usepackage{multirow}
\usepackage{rotating}
\usepackage{booktabs}
\usepackage{array}
\usepackage[ruled]{algorithm2e}

\newcommand{\shortname}{MSD-Mixer}
\newcommand{\longname}{\textbf{M}ulti-\textbf{S}cale \textbf{D}ecomposition MLP-\textbf{Mixer}}

\newcommand{\hl}[1]{#1}             

\begin{document}

\title{A Multi-Scale Decomposition MLP-Mixer for Time Series Analysis}

\author{Shuhan Zhong}
\orcid{0000-0003-4037-4288}
\affiliation{
  \department{Department of Computer Science and Engineering}
  \institution{The Hong Kong University of Science and Technology}
  }
\email{szhongaj@cse.ust.hk}

\author{Sizhe Song}
\orcid{0000-0001-8344-830X}
\affiliation{
  \department{Department of Computer Science and Engineering}
  \institution{The Hong Kong University of Science and Technology}
  }
\email{ssongad@cse.ust.hk}

\author{Weipeng Zhuo}
\orcid{0000-0002-1810-7071}
\affiliation{
  \department{Guangdong Provincial Key Laboratory IRADS and Department of Computer Science}
  \institution{BNU-HKBU United International College}
  }
\email{weipengzhuo@uic.edu.cn}

\author{Guanyao Li}
\orcid{0000-0002-3950-9360}
\affiliation{
  \department{Guangdong Enterprise Key Laboratory for Urban Sensing, Monitoring and Early Warning}
  \institution{Guangzhou Urban Planning and Design Survey Research Institute}
  }
\email{gyli@gzpi.com.cn}

\author{Yang Liu}
\orcid{0009-0005-8049-055X}
\affiliation{
  \department{Guangdong Enterprise Key Laboratory for Urban Sensing, Monitoring and Early Warning}
  \institution{Guangzhou Urban Planning and Design Survey Research Institute}
  }
\email{liuyang@gzpi.com.cn}

\author{S.-H. Gary Chan}
\orcid{0000-0003-4207-764X}
\affiliation{
  \department{Department of Computer Science and Engineering}
  \institution{The Hong Kong University of Science and Technology}
  }
\email{gchan@cse.ust.hk}

\begin{abstract}
Time series data, including univariate and multivariate ones, are characterized by unique composition and complex multi-scale temporal variations. They often require special consideration of decomposition and multi-scale modeling to analyze. Existing deep learning methods on this best fit to univariate time series only, and have not sufficiently considered sub-series modeling and decomposition completeness. \hl{To address these challenges, we propose \shortname{}, a \longname{}, which learns to explicitly decompose and represent the input time series in its different layers. To handle the multi-scale temporal patterns and multivariate dependencies, we propose a novel temporal patching approach to model the time series as multi-scale patches, and employ MLPs to capture intra- and inter-patch variations and channel-wise correlations. In addition, we propose a novel loss function to constrain both the mean and the autocorrelation of the decomposition residual for better decomposition completeness. Through extensive experiments on various real-world datasets for five common time series analysis tasks, we demonstrate that \shortname{} consistently and significantly outperforms other state-of-the-art algorithms with better efficiency.}
\end{abstract}

\maketitle

\pagestyle{\vldbpagestyle}
\begingroup\small\noindent\raggedright\textbf{PVLDB Reference Format:}\\
\vldbauthors. \vldbtitle. PVLDB, \vldbvolume(\vldbissue): \vldbpages, \vldbyear.\\
\href{https://doi.org/\vldbdoi}{doi:\vldbdoi}
\endgroup
\begingroup
\renewcommand\thefootnote{}\footnote{\noindent
This work is licensed under the Creative Commons BY-NC-ND 4.0 International License. Visit \url{https://creativecommons.org/licenses/by-nc-nd/4.0/} to view a copy of this license. For any use beyond those covered by this license, obtain permission by emailing \href{mailto:info@vldb.org}{info@vldb.org}. Copyright is held by the owner/author(s). Publication rights licensed to the VLDB Endowment. \\
\raggedright Proceedings of the VLDB Endowment, Vol. \vldbvolume, No. \vldbissue\ %
ISSN 2150-8097. \\
\href{https://doi.org/\vldbdoi}{doi:\vldbdoi} \\
}\addtocounter{footnote}{-1}\endgroup

\ifdefempty{\vldbavailabilityurl}{}{
\vspace{.3cm}
\begingroup\small\noindent\raggedright\textbf{PVLDB Artifact Availability:}\\
The source code, data, and/or other artifacts have been made available at \url{\vldbavailabilityurl}.
\endgroup
}

\section{Introduction}
\label{sec:intro}
A time series is a sequence of data points indexed in time order. It typically consists of successive numerical observations obtained in a fixed time interval. In multivariate time series, each observation contains more than one variable, which forms the "channel" dimension. With the fast development of sensing and data storage technologies, time series data is now becoming omnipresent in our lives, from weather conditions and urban traffic flow to personal health data monitored by smart wearable devices. \hl{The analysis of time series data, such as forecasting \cite{cui2021metro,tran2020deeptrans,bose2017probabilistic,fang2021mdtp}, missing data imputation \cite{khayati2020mind,miao2021generative,bansal2021missing}, anomaly detection \cite{ren2019time,schmidl2022anomaly,paparrizos2022volume,paparrizos2022tsb}, and classification \cite{cheng2023formertime,chowdhury2022tarnet,lee2008traclass}, is therefore facilitating more and more real-world applications, and attracts increasing research interest from both the academia and the industry.}

\begin{figure}
  \centering
  \includegraphics[width=0.85\linewidth]{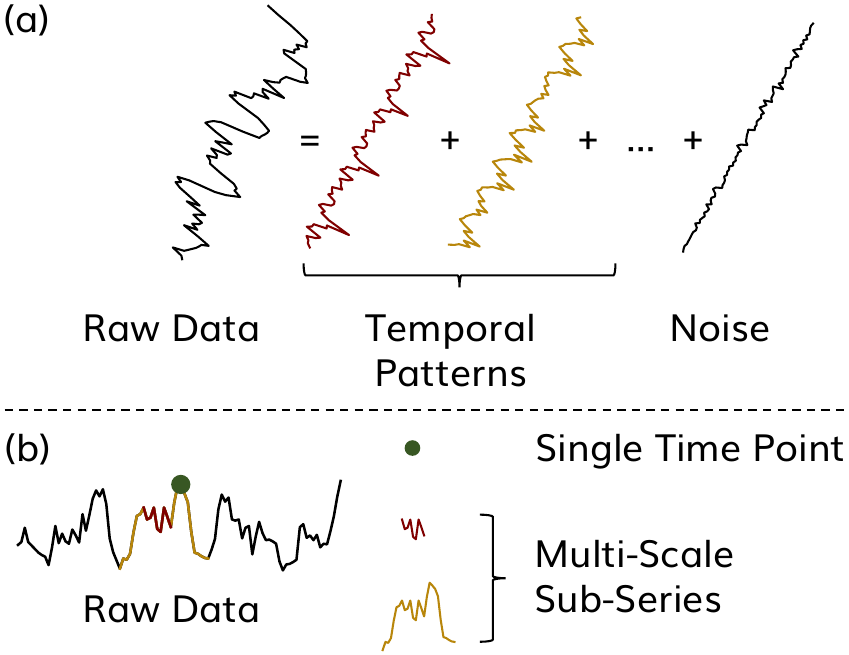}
  \caption{(a) Decomposition of time series. (b) Comparison of single time points and multi-scale sub-series.}
  \label{fig:intro}
\end{figure}

In contrast to images and natural languages, time series data is characterized by its special composition and complex temporal patterns or correlations. Specifically, each data point in a time series is actually a superposition of various underlying temporal patterns plus noise at that time (Figure \ref{fig:intro}(a)). To better model and analyze the data, it is hence important to decompose the data into disentangled components corresponding to different temporal patterns \cite{west1997time,hyndman2018forecasting}. Furthermore, time series data carries the semantic information of the temporal patterns in local consecutive data points termed \emph{sub-series}, rather than individual data point \cite{wu2023timesnet,nie2023time,wang2023micn,zeng2023transformers}. The temporal patterns are usually in multiple timescales, thus making it important to extract sub-series features and model their changes in multiple time scales (Figure \ref{fig:intro}(b)). To make the problem even more complex, multivariate time series may involve intricate correlations between different channels \cite{zhang2023crossformer,han2023capacity}. These characteristics all make time series analysis a challenging problem.

Recent approaches for time series analysis take advantage of the strong expressiveness of deep learning, especially the Transformer architecture \cite{vaswani2017attention}, and have achieved significant performance in various tasks \cite{wu2021autoformer,zhou2021informer,kitaev2020reformer,zhou2022fedformer}. \hl{However, it is recently pointed out that the Transformer is actually no better than multi-layer perceptrons (MLPs) in time series modeling, since Transformers are designed to embed information of single time points and model their pair-wise correlations, while time series data carries information in its multi-scale sub-series instead of single time points. At the same time, most approaches consider no or merely simple decomposition of temporal patterns \cite{wu2021autoformer,zhou2022fedformer}, which makes it hard for them to handle various intricate temporal patterns in time series data~\cite{wen2020fast,bandara2021mstl}}. Considering the composition of time series, some methods adopt a deep learning architecture that learns to decompose temporal patterns from the input for forecasting \cite{oreshkin2020n,challu2023nhits,woo2022etsformer}. However, without considering inter-channel dependency, \cite{oreshkin2020n} and \cite{challu2023nhits} are best applicable to univariate rather than multivariate time series. Furthermore, the aforementioned works did not investigate into the residual of the decomposition, which may lead to \emph{incomplete} decomposition, i.e., meaningful temporal patterns may be left in the residual and not utilized by the model.

In this work, we address these problems by proposing \shortname{}, a novel \longname{} to analyze both univariate and multivariate time series. \hl{\shortname{} is based exclusively on MLPs, which is simple but effective for time series modeling. To account for the special composition of time series data, \shortname{} explicitly decomposes the input time series into different components by generating their latent representations in different layers, and accomplishes the analysis task based on such representations. In \shortname{} we propose a novel \emph{multi-scale temporal patching} approach that divides the input time series into non-overlapping patches along the temporal dimension in each layer for sub-series modeling. Different layers have different patch sizes such that they can focus on different time scales. To better model multi-scale temporal patterns and inter-channel dependencies, \shortname{} employs MLPs along different dimensions to learn intra- and inter-patch variations as well as channel-wise correlations. In addition, to enhance the learning of the decomposition process, we propose a novel loss function to constrain both the mean and the autocorrelation of the decomposition residual during training. Used together with the loss function from the target analysis task, it helps \shortname{} to decompose the time series data more thoroughly for better analysis results.}

Empowered by the above-mentioned decomposition and multi-scale modeling features, \shortname{} distinguishes itself as a \emph{task-general} backbone that can be adapted for various time series analysis tasks. Through extensive experiments on various real-world datasets, we demonstrate that \shortname{} consistently outperforms both task-general and task-specific state-of-the-art approaches by a wide margin across five common time series analysis tasks, namely long-term forecasting (up to 9.8\% in MSE), short-term forecasting (up to 5.6\% in OWA), imputation (up to 46.1\% in MSE), anomaly detection (up to 33.1\% in F1-score) and classification (up to 36.3\% in Mean Rank).

To summarize, we make the following contributions in this paper:
\begin{itemize}
    \item \emph{A novel task-general backbone \shortname{}} that is well designed to analyze time series data by learning to explicitly decompose and represent the temporal patterns.
    \item \emph{A multi-scale temporal patching approach} in \shortname{} that facilitates modeling the time series data as multi-scale patches with MLPs, to better account for the multi-scale temporal patterns in the data.
    \item \emph{A residual loss} for \shortname{} to constrain both the mean and the autocorrelation of the decomposition residual for better decomposition completeness.
    \item \emph{Extensive experiments} on 26 datasets for five common time series analysis tasks to validate the effectiveness of \shortname{}.
\end{itemize}

The remainder of this paper is organized as follows: We first review related works in Section~\ref{sec:rw}, and elaborate on MSD-Mixer and its modules in Section~\ref{sec:msdmixer}. We show the experimental results in Section~\ref{sec:exp} and conclude in Section~\ref{sec:conclu}.

\section{Related Works}
\label{sec:rw}
\subsection{Classical Methods}
As one of the fundamental data modalities, time series has been well studied for long in various science and engineering domains that rely on temporal measurements, and has been mostly discussed for forecasting \cite{hyndman2018forecasting,faloutsos2018forecasting}. Regarding the special composition and the complex temporal patterns of time series data, early approaches employ manually designed rules or function models to decompose the time series data, such that the temporal patterns can be disentangled and modeled separately \cite{dagum2016seasonal,cleveland1990stl,dokumentov2015str,theodosiou2011forecasting,winters1960forecasting,holt2004forecasting,wen2019robuststl}. The decomposition usually consists of several components representing different temporal patterns, plus a residual which is supposed to be noise with no useful information. These approaches usually require considerable domain knowledge and manual effort to be adapted to specific domains, and are less expressive and scalable considering nowadays large multivariate time series datasets with complex temporal patterns and channel-wise correlations.

\subsection{Deep Models without Decomposition}
\hl{In recent years, deep learning has been widely applied in time series analysis for its strong expressiveness and scalability on large and complex datasets. The deep learning based approaches either apply MLP \cite{zhang2022less}, convolutional neural network (CNN) \cite{ismail2020inceptiontime,liu2022scinet,wu2023timesnet}, recurrent neural network (RNN) \cite{bai2018empirical}, Transformer \cite{cheng2023formertime,nie2023time,xu2022anomaly}, or their combination \cite{karim2019multivariate,wang2023micn,bansal2021missing} to model the time series data for specific tasks.
Among them, RNNs have been pointed out for their deficiency in modeling long sequences which are common in time series analysis tasks. Its difficulty with parallelized training also greatly affects their efficiency. CNNs, instead, usually require special attention to make the trade-off between the number of layers and the effective receptive fields, or consider the design of dilation or pooling rate when applied for time series analysis \cite{li2021modeling,lea2017temporal}.
Transformer-based models are taking the lead in many time series analysis tasks due to the powerful capability of attention mechanism to capture long-sequence dependencies. Many works have been done to further improve the efficiency \cite{kitaev2020reformer,zhou2021informer} and effectiveness \cite{wu2021autoformer,zhou2022fedformer,shabani2023scaleformer} of the Transformer for time series data. However, it is recently shown that the Transformer, which relies on point embedding and their pair-wise correlations, is not a promising choice for time series data, since the semantic information is embedded in the sub-series level variations instead of single time points \cite{zeng2023transformers}. In light of this, PatchTST \cite{nie2023time} and TimesNet \cite{wu2023timesnet} are proposed to combine patch modeling with Transformer and CNN for time series data.}
Despite the above-mentioned achievements, most deep learning based approaches do not consider the decomposition of temporal patterns, or only simply consider the decomposition of very limited types and number of components \cite{wu2021autoformer,zhou2022fedformer}, which makes it hard for them to deal with multiple intricate temporal patterns in time series data~\cite{bandara2021mstl}.

\subsection{Deep Models with Decomposition}
By combining deep learning with decomposition, N-BEATS~\cite{oreshkin2020n}, N-HiTS~\cite{challu2023nhits}, and ETSformer~\cite{woo2022etsformer} show satisfactory results in time series forecasting. However, N-BEATS and N-HiTS do not consider the inter-channel correlation, which has been shown critical in multivariate time series analysis tasks. In addition, they are based on plain MLP on the temporal dimension while ETSformer~\cite{woo2022etsformer} is based on self-attention for temporal modeling, all of which do not take into account the sub-series level features. Furthermore, they simply ignore the residual of the decomposition, which may lead to incomplete decomposition that meaningful temporal patterns can be left in the residual and not utilized. Besides, all these schemes have only been tested on the forecasting task, leaving other analysis tasks such as imputation, anomaly detection, and classification unexplored.

In comparison, we propose \shortname{} that advances them with \emph{multi-scale temporal patching} and multi-dimensional MLP mixing for multi-scale sub-series and inter-channel modeling. Meanwhile, we propose a \emph{residual loss} for better completeness of the decomposition process in \shortname{}. Furthermore, we experiment \shortname{} and compare it with state-of-the-art methods on various datasets across the forecasting, imputation, anomaly detection, and classification tasks to show its superior modeling ability over other methods.

\section{\shortname{}}
\label{sec:msdmixer}
In this section, we first formally introduce the definition of general time series analysis problems and time series decomposition in \ref{sec:prob}. Then, we overview the architecture and workflow of our proposed \shortname{} in Section \ref{sec:overview}, followed by elaboration on the key designs in \shortname{} in Sections \ref{sec:patching} to \ref{sec:resloss}, then summarize in Section \ref{sec:sum}.

\subsection{Problem Settings}
\label{sec:prob}
\subsubsection{Time Series Analysis}
In this paper, we summarize the general learning-based time series analysis tasks, including but not limited to \emph{forecasting}, \emph{imputation}, \emph{anomaly detection}, and \emph{classification}, as the following problem: Given a dataset $\mathcal{D}$ containing sample pairs $(\boldsymbol{X},\boldsymbol{Y})$, where $\boldsymbol{X}\in\mathbb{R}^{C\times L}$ denotes the input multivariate time series with $C$ channels and $L$ time steps, and $\boldsymbol{Y}$ denotes the label whose form is subject to the target time series analysis task. The time series analysis problem is to obtain an optimal function $\mathcal{F}(\cdot)$ on the dataset that maps the input $\boldsymbol{X}$ to its corresponding label as $\boldsymbol{\hat{Y}}=\mathcal{F}(\boldsymbol{X})$, such that the difference between the prediction $\boldsymbol{\hat{Y}}$ and the ground truth $\boldsymbol{Y}$ is minimized.

Different target time series analysis tasks have different forms of $\boldsymbol{Y}$ and $\boldsymbol{\hat{Y}}$. For example, $\boldsymbol{Y}\in\mathbb{R}^{C\times H}$ in a forecasting task with horizon size $H$, and $\boldsymbol{Y}\in\mathbb{R}^{M}$ in a classification task with $M$ target classes. Different tasks also employ different metrics to measure the difference between $\boldsymbol{Y}$ and $\boldsymbol{\hat{Y}}$ for computing the task-specific loss, e.g., cross-entropy loss used in classification and mean square error loss used in forecasting.

\subsubsection{Time Series Analysis with Decomposition}
Existing deep learning approaches for time series generally learn to directly represent the input $\boldsymbol{X}$ and map it to the output, which makes it hard for them to handle multiple intricate temporal patterns. Instead, we consider the decomposition of $\boldsymbol{X}$ as:
\begin{equation}
    \label{eq:decomp}
    \boldsymbol{X}=\sum_{i=1}^{k}\boldsymbol{S}_i+\boldsymbol{R},
\end{equation}
where $\boldsymbol{S}_i, \boldsymbol{R}\in\mathbb{R}^{C\times L}$ denote the $i$-th component ($i=1,...,k$) and the residual, respectively. Suppose each component $\boldsymbol{S}_i$ has a lower-dimensional representation $\boldsymbol{E}_i$, then the target $\mathcal{F}$ can be divided and conquered by a set of functions $f_i(\cdot)$ of the component representations:
\begin{equation}
    \label{eq:tgt}
    \boldsymbol{\hat{Y}}=\mathcal{F}(\boldsymbol{X})=\sum_{i=1}^{k}f_i(\boldsymbol{E}_i).
\end{equation}

\begin{figure}[]
  \centering
  \includegraphics[width=\linewidth]{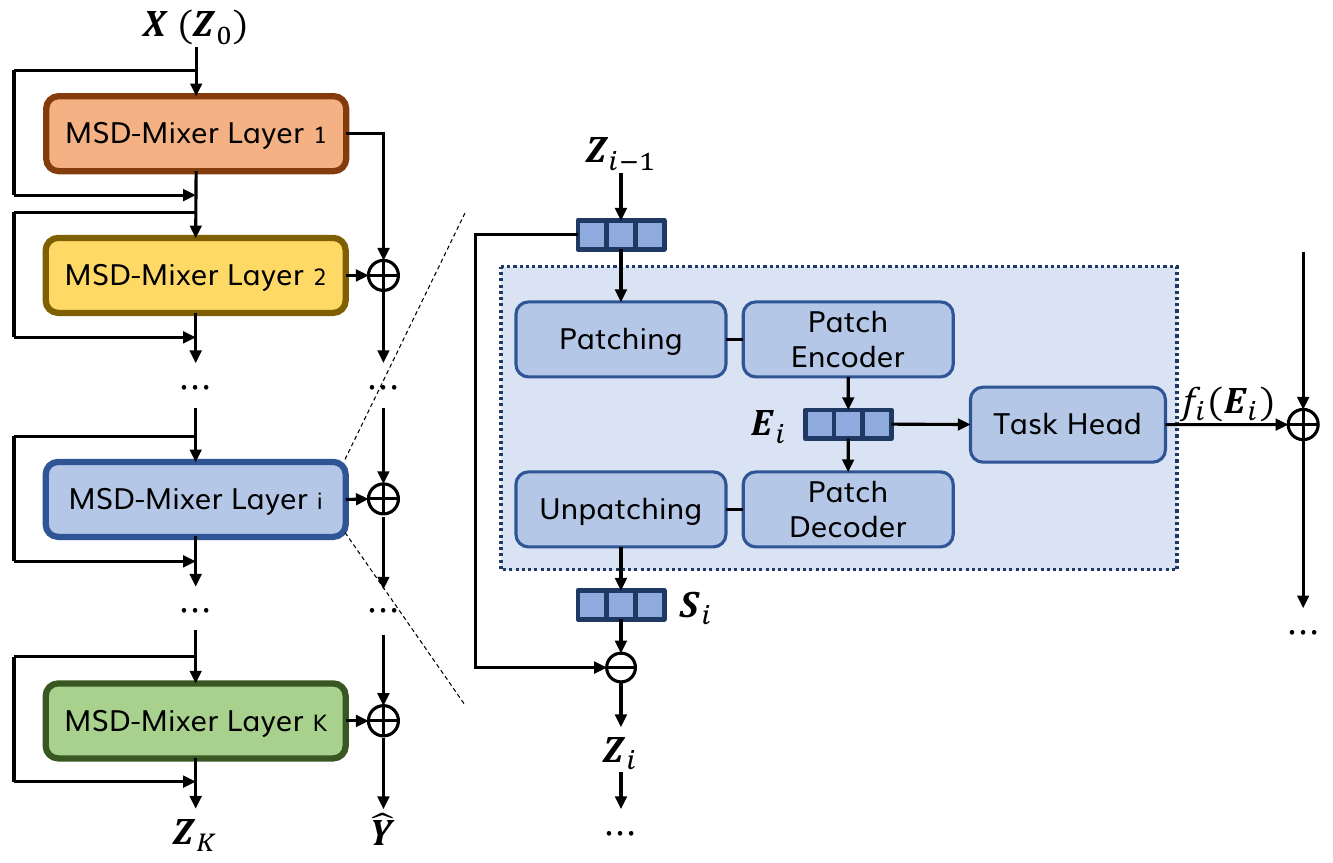}
  \caption{\shortname{} overview.}
  \label{fig:overview}
\end{figure}

\subsection{\shortname{} Overview}
\label{sec:overview}
Figure \ref{fig:overview} shows the overall architecture of \shortname{}. \shortname{} comprises a stack of $k$ layers, and learns to hierarchically decompose the input $\boldsymbol{X}$ into $k$ components $\{\boldsymbol{S}_1,...,\boldsymbol{S}_k\}$ by generating their lower-dimensional representations $\{\boldsymbol{E}_1,...,\boldsymbol{E}_k\}$ in the corresponding layers. The number of layers and components $k$ is a hyperparameter in \shortname{} which should be determined according to the properties of the dataset. Here we define $\boldsymbol{Z}_0=\boldsymbol{X}$, and
\begin{equation}
    \label{eq:z}
    \boldsymbol{Z}_i=\boldsymbol{X}-\sum_{j=1}^{i}\boldsymbol{S}_j, (i=1,...,k),
\end{equation}
such that $\boldsymbol{Z}_i$ specifies the remaining part after the first $i$ components has been decomposed from the input $\boldsymbol{X}$, and we have
\begin{equation}
    \label{eq:z_itr}
    \boldsymbol{Z}_i=\boldsymbol{Z}_{i-1}-\boldsymbol{S}_i.
\end{equation}
As is shown in Figure \ref{fig:overview}, the $i$-th layer of \shortname{} takes the remaining part $\boldsymbol{Z}_{i-1}$ from the previous layer as input, and learns to represent $\boldsymbol{Z}_{i-1}$ with a lower dimensional representation $\boldsymbol{E}_i$. The represented part is the $i$-th component $\boldsymbol{S}_i$. More specifically, within each layer, $\boldsymbol{Z}_{i-1}$ is first patched in the Patching module, and then fed into the Patch Encoder module to generate the representation of $i$-th components as $\boldsymbol{E}_{i}=g_i(\boldsymbol{Z}_{i-1})$. The Patch Decoder module then reconstructs $\boldsymbol{S}_i$ from $\boldsymbol{E}_{i}$, and the Unpatching module unpatches it to the original dimensionality. After that, $\boldsymbol{S}_i$ is subtracted from $\boldsymbol{Z}_{i-1}$ to obtain $\boldsymbol{Z}_{i}$. $\boldsymbol{Z}_{i}$ is fed to the next layer as input for further decomposition, and $\boldsymbol{E}_{i}$ is projected by linear layers and summed to obtain $\boldsymbol{Y}$ following Equation \ref{eq:tgt}.

\begin{figure}[]
  \centering
  \includegraphics[width=0.85\linewidth]{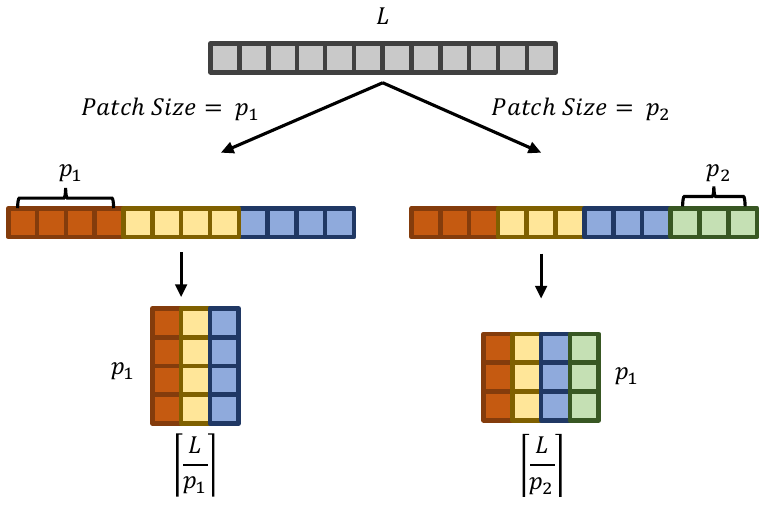}
  \caption{Examples of multi-scale temporal patching. The channel dimension is omitted for simplicity.}
  \label{fig:patching}
\end{figure}

\subsection{Multi-Scale Temporal Patching}
\label{sec:patching}
Considering the importance of multi-scale sub-series modeling for time series analysis, we introduce \emph{multi-scale temporal patching} in \shortname{} such that different layers can focus on different sub-series features. Each layer of \shortname{} has a predefined patch size $p_i$, which is a hyperparameter to be determined or tuned for specific datasets.

We depict the patching process in Figure \ref{fig:patching}. To transform an input time series with $C$ channels and $L$ time steps into patches with patch size $p$, we first pad the time series with zeros at the beginning of the time series to ensure the length is divisible by $p$, and then segment the time series along the temporal dimension into non-overlapping patches with stride $p$. We then permute the data to create a new dimension for the patches, resulting in a high dimensional tensor of $C\times L'\times p$, where $L'=\lceil L/p\rceil$

\begin{figure*}
  \centering
  \includegraphics[width=0.9\linewidth]{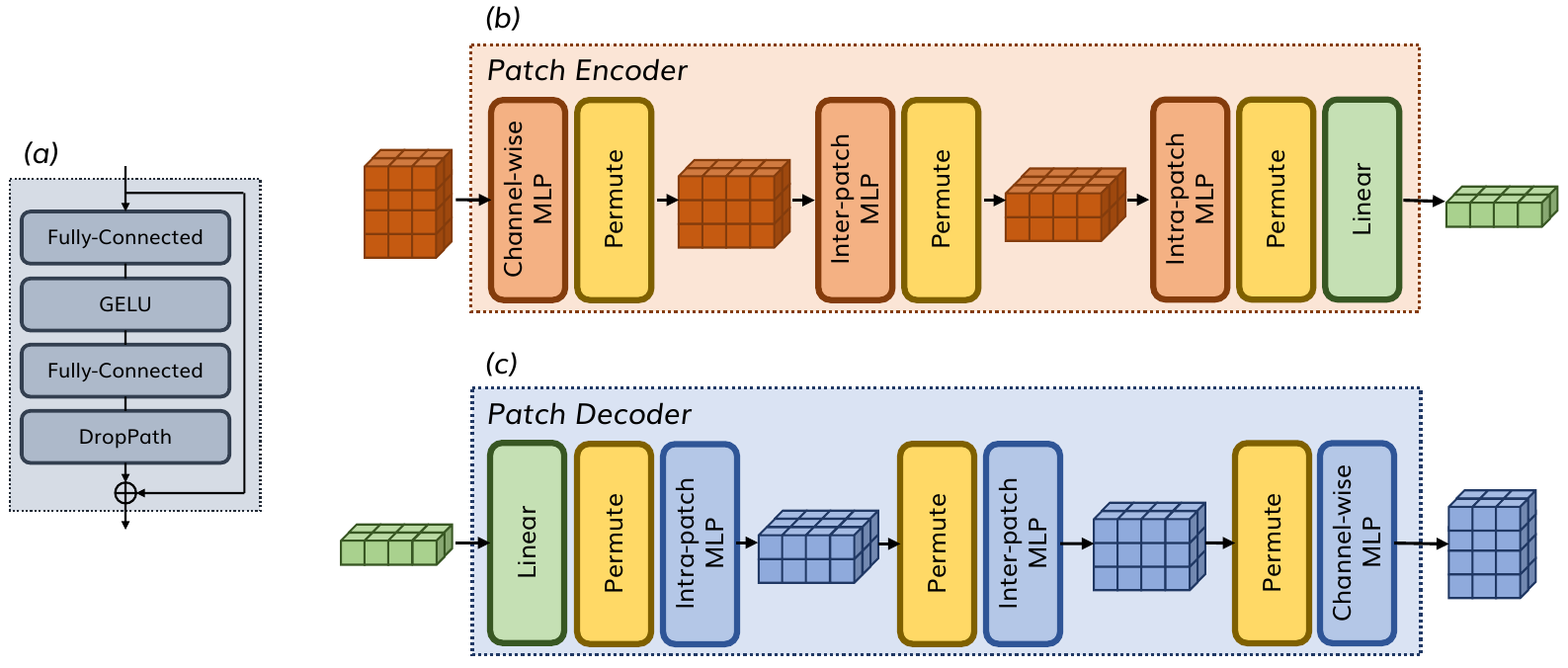}
  \caption{(a) MLP block. (b) Patch Encoder. (c) Patch Decoder.}
  \label{fig:mlpencdec}
\end{figure*}
\subsection{Patch Encoder and Decoder}
The Patch Encoder and Decoder modules are based exclusively on MLPs along different dimensions for feature extraction. We show the design of each MLP block in Figure \ref{fig:mlpencdec}(a), which simply consists of two fully connected layers, a GELU nonlinearity layer, and a DropPath layer \cite{larsson2017fractalnet}, together with a residual connection that adds the input to the output. We use the following three types of MLP blocks in Patch Encoder and Decoder modules:
\begin{itemize}
    \item \emph{The channel-wise MLP block} allows communication between different channels, to capture inter-channel correlations.
    \item \emph{The inter-patch MLP block} allows communication between different patches, to capture global contexts.
    \item \emph{The intra-patch MLP block} allows communication between different time steps within a patch, to capture sub-series level variations.
\end{itemize}
As shown in Figure \ref{fig:mlpencdec}(b) the Patch Encoder module consists of a channel-wise MLP block, an inter-patch MLP block, an intra-patch MLP block, and a linear layer in order to produce the component representation $\boldsymbol{E}_{i}$ from the patched $\boldsymbol{Z}_{i-1}$. The Patch Decoder module (\ref{fig:mlpencdec}(c)) consists of the same number and type of blocks as the Patch Encoder module, but in a reversed order to reconstruct $\boldsymbol{S}_i$ from $\boldsymbol{E}_{i}$.

\subsection{Residual Loss}
\label{sec:resloss}
\hl{
The residual of the decomposition is useful in checking whether the information in the data has been adequately captured into the components. An ideal decomposition should yield a residual with the following two properties:
\begin{itemize}
  \item It should have zero mean. If the residual has a mean other than zero, then there can be biases left in the residual.
  \item It should contain no autocorrelation. The stronger the autocorrelation is in the residual, the more likely there can be temporal patterns such as trends and periodic information left in the residual.
\end{itemize}
A residual that does not satisfy these properties indicates the incompleteness of the decomposition, which means useful information has not been fully accounted for by the components.}

\hl{
By jointly considering the two properties of the decomposition residual, we propose a novel \emph{residual loss} to train \shortname{} such that it can learn to achieve better decomposition completeness. The \emph{residual loss} minimizes both the mean and the autocorrelation of the residual. The autocorrelation of the residual can be measured by its autocorrelation coefficients which are defined in $[-1,1]$. A larger absolute value of the coefficient indicates a stronger correlation. It is usually expected that the autocorrelation coefficients of a successful decomposition should lie within $\pm2/\sqrt{L}$ where $L$ is the series length.} In a \shortname{} with $k$ layers, $\boldsymbol{Z}_{k}$ output by the last layer specifies the residual of the decomposition. We first compute the autocorrelation coefficient matrix $\boldsymbol{A}=\{a_{i,j}\}\in \mathbb{R}^{C\times (L-1)}$ of $\boldsymbol{Z}_{k}$ as:
\begin{equation}
    \label{eq:acf}
    a_{i,j}=\frac{\sum_{t=j+1}^{L}(z_{i,t}-\bar{z_i})(z_{i,t-j}-\bar{z_i})}{\sum_{t=1}^{L}(z_{i,t}-\bar{z_i})^2},
\end{equation}
where $z_{i,j}$ is the $i$-th channel and $j$-th time step of $\boldsymbol{Z}_{k}$. We then define the \emph{residual loss} by:
\begin{equation}
    \label{eq:acf_loss}
    \mathcal{L}_r=\frac{\sum_{i,j}z_{i,j}^2}{C\times L}+\frac{\sum_{i,j}(\mathrm{ReLU}(|a_{i,j}|-\alpha/\sqrt{L}))^2}{C\times(L-1)}.
\end{equation}
The first term of $\mathcal{L}_r$ minimizes the mean of the residual. And the right term imposes a constraint on the autocorrelation coefficients of the residual, where $\alpha$ is a hyperparameter controlling the maximum tolerance of the autocorrelation coefficients. We finally train \shortname{} by simultaneously optimizing the weighted sum of the task-specific loss and the \emph{residual loss}:
\begin{equation}
    \label{eq:loss}
    \mathcal{L}=\mathcal{L}_t+\lambda\mathcal{L}_r.
\end{equation}

\begin{algorithm}
\small
\caption{Training of \shortname{}.}
\label{algo:msd-mixer}
\textbf{Input:} Training set $\mathcal{D}=\{(\boldsymbol{X},\boldsymbol{Y})\}$, number of layers $k$, patch size for each layer $p_1,\cdots,p_k$.\\
\textbf{Output:} Trained \shortname{}.\\
\nl Initialize \shortname{} with $k$ layers of patch size $p_1,\cdots,p_k$.\\
\nl \Repeat{convergence}{
  \nl Sample ($\boldsymbol{X}$,$\boldsymbol{Y}$) from $\mathcal{D}$.\\
  \nl $\boldsymbol{Z}_0=\boldsymbol{X}$.\\
	\nl	\For{$i= 1, 2, \ldots, k$}{
    \nl Patch $\boldsymbol{Z}_{i-1}$ with $p_i$.\\
    \nl Compute $\boldsymbol{E}_i = g_i(\boldsymbol{Z}_{i-1})$ with $i$-th layer's Patch Encoder.\\
    \nl Compute $\boldsymbol{S}_i = h_i(\boldsymbol{E}_i)$ with $i$-th layer's Patch Decoder.\\
    \nl Unpatch $\boldsymbol{S}_i$ with $p_i$.\\
    \nl $\boldsymbol{Z}_i=\boldsymbol{Z}_{i-1}-\boldsymbol{S}_i$.
		}
	\nl Compute $\boldsymbol{\hat{Y}}=\sum_{i=1}^k f_i(\boldsymbol{E}_i)$ with Task Head modules.\\
	\nl Computer loss according to Equation~\ref{eq:acf}--\ref{eq:loss}.\\
	\nl Back propagation.
	}
\nl \Return{Trained \shortname{}.}
\end{algorithm}
\subsection{Summary}
\label{sec:sum}
We summarize the overall training process of \shortname{} in Algorithm~\ref{algo:msd-mixer}. Given a training dataset $\mathcal{D}$ with time series data and corresponding labels as $\mathcal{D}=\{(\boldsymbol{X},\boldsymbol{Y})\}$, we train \shortname{} based on the dataset until the loss converges. During the process, we first sample $(\boldsymbol{X},\boldsymbol{Y})$ from $\mathcal{D}$ (line 3), and then initialize $\boldsymbol{Z}_0$ to be $\boldsymbol{X}$ (line 4). After that, for each layer $i \in [1,k]$, we patch $\boldsymbol{Z}_{i-1}$ using the patch size $p_i$ (line 6). The patched input is then fed into MLP-Mixer to generate $\boldsymbol{E}_i$ (line 7). This learned representation $\boldsymbol{E}_i$ is eventually decoded and unpatched to reconstruct the input in each layer (lines 8 -- 10). The labels are predicted with loss computed for back propagation (lines 11 -- 13). This whole process is repeated until convergence to return the trained MSD-Mixer model (line 14).

\section{Illustrative Experimental Results}
\label{sec:exp}
\hl{In this section, we first overview the experiment setup and key results in Section \ref{sec:exp_overview}. Then, in Section \ref{sec:ltf} to \ref{sec:ad}, we discuss in detail the experiments and results of different time series analysis tasks, followed by ablation studies on our proposed modules in Section \ref{sec:abl}. Lastly, we study the efficiency of \shortname{} by comparing the number of model parameters and training time consumption of different approaches in Section \ref{sec:eff}, and empirically analyze the decomposition of \shortname{} by example cases in Section \ref{sec:cs}.}

\begin{table}
  \small
  \renewcommand\arraystretch{0.9}
  \centering
  \caption{Summary of Tasks, Datasets and Metrics}
    \begin{tabular}{>{\raggedright\arraybackslash}m{0.6in}|>{\raggedright\arraybackslash}m{0.8in}|>{\raggedright\arraybackslash}m{1.4in}}
    \toprule
    Tasks & Datasets \cite{wu2023timesnet} & Metrics \\
    \midrule
    Long-Term Forecasting & ETT (4 subsets), ECL, Weather, Traffic, Exchange & Mean Square Error (MSE), Mean Absolute Error (MAE) \\
    \midrule
    Short-Term Forecasting & M4 (6 subsets) & SMAPE, MASE, OWA \cite{makridakis2018m4} \\
    \midrule
    Imputation & ETT (4 subsets), ECL, Weather & MSE, MAE \\
    \midrule
    Anomaly Detection & SMD, MSL, SMAP, SWaT, PSM & F1-Score \\
    \midrule
    Classification & UEA (10 subsets \cite{cheng2023formertime}) & Accuracy \\
    \bottomrule
    \end{tabular}%
  \label{tab:task_summary}%
\end{table}%

\begin{table*}[]
  \small
  \centering
  \setlength\tabcolsep{3pt}
  \renewcommand\arraystretch{0.8}
  \caption{\hl{Overall performance comparison with task-general baselines. Each number of a scheme in the table represents in how many benchmarks the scheme performs the best. The best results are in \textbf{bold} and the second bests are \underline{underlined}.}}
    \begin{tabular}{c|c|c|c|c|c|c|c|c|c|c}
    \toprule
    \multirow{2}{*}{Task}  & \multirow{2}{*}{\# of Benchmarks} & MSD-Mixer    & PatchTST      & Crossformer   & TimesNet       & DLinear & ETSformer & NST    & FEDformer & LightTS \\
                           &                                   & (Ours)       & (2023)        & (2023)        & (2023)         & (2023)  & (2022)    & (2022) & (2022)    & (2022)  \\
    \midrule                                                                                                                                                                                                                       
    Long-Term Forecasting  & 64                                & \textbf{45}  & 7             & \underline{8} & 1              & 3       & 2         & 0      & 1         & 1       \\
    Short-Term Forecasting & 15                                & \textbf{15}  & 0             & 0             & 0              & 0       & 0         & 0      & 0         & 0       \\
    Imputation             & 48                                & \textbf{45}  & 0             & 0             & \underline{9}  & 0       & 0         & 0      & 0         & 0       \\
    Anomaly Detection      & 5                                 & \textbf{4}   & 0             & 0             & 1              & 0       & 0         & 0      & 0         & 0       \\
    Classification         & 10                                & \textbf{5}   & 0             & 0             & 0              & 0       & 0         & 0      & 0         & 0       \\
    \midrule                                                                                                                                                                                                                         
    Total                  & 142                               & \textbf{114} & 7             & 8             & \underline{11} & 3       & 2         & 0      & 1         & 1       \\
    \bottomrule
    \end{tabular}%
  \label{tab:1st}%
\end{table*}%
\subsection{Overview}
\label{sec:exp_overview}

In order to validate the modeling ability of \shortname{}, we conduct extensive experiments on a wide range of well-adopted benchmark datasets across \emph{five} most common time series analysis tasks, including long-term forecasting, short-term forecasting, imputation, anomaly detection, and classification. The tasks and benchmark datasets are of different characteristics that we leverage them to investigate on different aspects of \shortname{}. Table~\ref{tab:task_summary} summarizes the five tasks, datasets, and evaluation metrics we use in the experiments.

\subsubsection{Baselines}
We compare our proposed \shortname{} with state-of-the-art \emph{task-general} approaches that can serve as general solutions to various time series analysis tasks, as well as \emph{task-specific} approaches that are proposed for specific time series analysis tasks. 

For \emph{task-general} baselines, we select approaches that cover mainstream deep learning architectures, including CNN, Transformer, and MLP. \hl{Among them, TimesNet~\cite{wu2023timesnet} and PatchTST~\cite{nie2023time} combine sub-series modeling with CNN and Transformer, respectively. Crossformer~\cite{zhang2023crossformer} is a Transformer-based approach with special designs for channel-wise correlations. ETSformer~\cite{woo2022etsformer} is among the first to leverage Transformer for decomposition. NST~\cite{liu2022stationary} and FEDformer~\cite{zhou2022fedformer} are also Transformer-based approaches that consider the stationarity and frequency domain features of time series data. DLinear~\cite{zeng2023transformers} and LightTS~\cite{zhang2022less} are MLP-based light-weight approaches. LightTS further considers channel-wise and local-global features in the data.}

In addition, we introduce and compare with extra state-of-the-art task-specific approaches for tasks in the corresponding sections, including long-term forecasting (Section~\ref{sec:ltf}), short-term forecasting (Section~\ref{sec:stf}), anomaly detection (Section~\ref{sec:ad}), and classification (Section~\ref{sec:cls}).

\subsubsection{Implementation}
We follow the implementation of baselines in~\cite{wu2023timesnet}. All models including \shortname{} and the baselines are implemented with PyTorch and trained with a single NVIDIA GeForce RTX 4090 GPU with 24 GB memory. We search the best number of layers from 4 to 6, and dimensions of the model from 64 to 512 for \shortname{} with different datasets. We set the patch size in \shortname{} by considering the series length and sampling interval of the dataset. For instance, the ETTm1 dataset provides two years' data of electricity transformer temperature from two separate counties in China. It contains time series with a length of 96 samples and the sampling interval is 15 minutes, i.e., two samples are collected 15 minutes apart. To model the time series efficiently, we use five layers in \shortname{} with patch sizes for each layer as \{24, 12, 4, 2, 1\}, which correspond to the sub-series of 6 hours (15min $\times$ 24), 3 hours, 1 hour, 30 minutes, and 15 minutes, respectively.

\subsubsection{Overall Performance}
Table \ref{tab:1st} summarizes the overall performance of task-general schemes. \hl{As shown in the table, our proposed \shortname{} outperforms other state-of-the-art baselines significantly in all the benchmarks across the five tasks. Benefiting from the special design of the decomposition, multi-scale temporal patching, and the \emph{residual loss} components, \shortname{} is far ahead of its CNN-based (TimesNet), Transformer-based (PatchTST, Crossformer, ETSformer, NST, FEDformer) and MLP-based (DLinear, LightTS) task-general counterparts by a large margin, demonstrating its great and comprehensive modeling ability for time series analysis.}

\begin{table}
  \small
  \centering
  \renewcommand\arraystretch{0.8}
  \caption{Statistics of datasets for long-term forecasting.}
  \begin{tabular}{cccc}
    \toprule
    Dataset & Dim   & Total Timesteps & Frequency \\
    \midrule
    ETTm1, ETTm2 & 7     & 69680 & 15 mins \\
    ETTh1, ETTh2 & 7     & 17420 & 1 hour \\
    ECL   & 321   & 26304 & 10 mins \\
    Traffic & 862   & 17544 & 1 hour \\
    Weather & 21    & 52696 & 10 mins \\
    Exchange & 8     & 7588  & 1 day \\
    \bottomrule
  \end{tabular}%
  \label{tab:ltfdata}%
\end{table}%
\begin{table*}
  \small
  \setlength\tabcolsep{2.8pt}
  \renewcommand\arraystretch{0.85}
  \centering
  \caption{\hl{Long-term forecasting results. The best results are in \textbf{bold} and the second bests are \underline{underlined}. (*Task-specific baseline.)}}
    \begin{tabular}{cc|cc|cc|cc|cc|cc|cc|cc|cc|cc|cc}
    \toprule
    \multicolumn{2}{c|}{\multirow{2}{*}{Models}} & \multicolumn{2}{c|}{MSD-Mixer} & \multicolumn{2}{c|}{Scaleformer*} & \multicolumn{2}{c|}{PatchTST} & \multicolumn{2}{c|}{Crossformer} & \multicolumn{2}{c|}{TimesNet} & \multicolumn{2}{c|}{DLinear} & \multicolumn{2}{c|}{ETSformer} & \multicolumn{2}{c|}{NST}   & \multicolumn{2}{c|}{FEDformer} & \multicolumn{2}{c}{LightTS} \\
    \multicolumn{2}{c|}{}                        & \multicolumn{2}{c|}{(Ours)}    & \multicolumn{2}{c|}{(2023)}       & \multicolumn{2}{c|}{(2023)}   & \multicolumn{2}{c|}{(2023)}      & \multicolumn{2}{c|}{(2023)}   & \multicolumn{2}{c|}{(2023)}  & \multicolumn{2}{c|}{(2022)}    & \multicolumn{2}{c}{(2022)} & \multicolumn{2}{c|}{(2022)}    & \multicolumn{2}{c}{(2022)}  \\
    \midrule
    \multicolumn{2}{c|}{Metric} & MSE   & MAE   & MSE   & MAE   & MSE   & MAE   & MSE   & MAE   & MSE   & MAE   & MSE   & MAE   & MSE   & MAE   & MSE   & MAE   & MSE   & MAE  & MSE   & MAE \\
    \midrule
    \multicolumn{1}{c|}{\multirow{4}{*}{\begin{sideways}ETTm1\end{sideways}}}       & 96    & \textbf{0.304}    & \textbf{0.351}    & 0.392             & 0.415 & 0.334             & \underline{0.372} & \underline{0.316} & 0.373             & 0.338             & 0.375             & 0.345             & \underline{0.372} & 0.375          & 0.398          & 0.386             & 0.398             & 0.379             & 0.419 & 0.374          & 0.400             \\
    \multicolumn{1}{c|}{}                                                           & 192   & \textbf{0.344}    & \textbf{0.375}    & 0.437             & 0.451 & 0.378             & 0.394             & 0.377             & 0.411             & \underline{0.374} & \underline{0.387} & 0.380             & 0.389             & 0.408          & 0.410          & 0.459             & 0.444             & 0.426             & 0.441 & 0.400          & 0.407             \\
    \multicolumn{1}{c|}{}                                                           & 336   & \textbf{0.370}    & \textbf{0.395}    & 0.499             & 0.478 & \underline{0.406} & 0.414             & 0.431             & 0.442             & 0.410             & \underline{0.411} & 0.413             & 0.413             & 0.435          & 0.428          & 0.495             & 0.464             & 0.445             & 0.459 & 0.438          & 0.438             \\
    \multicolumn{1}{c|}{}                                                           & 720   & \textbf{0.427}    & \textbf{0.428}    & 0.584             & 0.536 & \underline{0.462} & \underline{0.445} & 0.600             & 0.547             & 0.478             & 0.450             & 0.474             & 0.453             & 0.499          & 0.462          & 0.585             & 0.516             & 0.543             & 0.490 & 0.527          & 0.502             \\
    \midrule                                                                                                                                                                                                                                                                                                                                                                                                                                                                                                          
    \multicolumn{1}{c|}{\multirow{4}{*}{\begin{sideways}ETTm2\end{sideways}}}       & 96    & \textbf{0.169}    & \textbf{0.259}    & 0.182             & 0.276 & \underline{0.175} & \textbf{0.259}    & 0.236             & 0.281             & 0.187             & 0.267             & 0.193             & 0.292             & 0.189          & 0.280          & 0.192             & 0.274             & 0.203             & 0.287 & 0.209          & 0.308             \\
    \multicolumn{1}{c|}{}                                                           & 192   & \textbf{0.232}    & \textbf{0.300}    & 0.252             & 0.319 & \underline{0.240} & \underline{0.302} & 0.294             & 0.349             & 0.249             & 0.309             & 0.284             & 0.362             & 0.253          & 0.319          & 0.280             & 0.339             & 0.269             & 0.328 & 0.311          & 0.382             \\
    \multicolumn{1}{c|}{}                                                           & 336   & \textbf{0.292}    & \textbf{0.337}    & 0.335             & 0.372 & \underline{0.302} & \underline{0.342} & 0.353             & 0.382             & 0.321             & 0.351             & 0.369             & 0.427             & 0.314          & 0.357          & 0.334             & 0.361             & 0.325             & 0.366 & 0.442          & 0.466             \\
    \multicolumn{1}{c|}{}                                                           & 720   & \textbf{0.392}    & \underline{0.398} & 0.460             & 0.446 & \underline{0.399} & \textbf{0.397}    & 0.588             & 0.547             & 0.408             & 0.403             & 0.554             & 0.522             & 0.414          & 0.413          & 0.417             & 0.413             & 0.421             & 0.415 & 0.675          & 0.587             \\
    \midrule                                                                                                                                                                                                                                                                                                                                                                                                                                                                                  
    \multicolumn{1}{c|}{\multirow{4}{*}{\begin{sideways}ETTh1\end{sideways}}}       & 96    & \underline{0.377} & \textbf{0.391}    & 0.404             & 0.441 & 0.444             & 0.438             & 0.386             & 0.429             & 0.384             & 0.402             & 0.386             & \underline{0.400} & 0.494          & 0.479          & 0.513             & 0.491             & \textbf{0.376}    & 0.419 & 0.424          & 0.432             \\
    \multicolumn{1}{c|}{}                                                           & 192   & 0.427             & \textbf{0.422}    & 0.438             & 0.461 & 0.488             & 0.463             & \textbf{0.419}    & 0.444             & 0.436             & \underline{0.429} & 0.437             & 0.432             & 0.538          & 0.504          & 0.534             & 0.504             & \underline{0.420} & 0.448 & 0.475          & 0.462             \\
    \multicolumn{1}{c|}{}                                                           & 336   & 0.469             & \textbf{0.443}    & 0.464             & 0.477 & 0.525             & 0.484             & \textbf{0.440}    & 0.461             & 0.491             & 0.469             & 0.481             & \underline{0.459} & 0.574          & 0.521          & 0.588             & 0.535             & \underline{0.459} & 0.465 & 0.518          & 0.488             \\
    \multicolumn{1}{c|}{}                                                           & 720   & \textbf{0.485}    & \textbf{0.475}    & 0.507             & 0.516 & 0.532             & 0.510             & 0.519             & 0.524             & 0.521             & \underline{0.500} & 0.519             & 0.516             & 0.562          & 0.535          & 0.643             & 0.616             & \underline{0.506} & 0.507 & 0.547          & 0.533             \\
    \midrule                                                                                                                                                                                                                                                                                                                                                                                                                                                                                  
    \multicolumn{1}{c|}{\multirow{4}{*}{\begin{sideways}ETTh2\end{sideways}}}       & 96    & \textbf{0.284}    & \textbf{0.345}    & 0.335             & 0.385 & \underline{0.312} & \underline{0.358} & 0.401             & 0.464             & 0.340             & 0.374             & 0.333             & 0.387             & 0.340          & 0.391          & 0.476             & 0.458             & 0.358             & 0.397 & 0.397          & 0.437             \\
    \multicolumn{1}{c|}{}                                                           & 192   & \textbf{0.362}    & \textbf{0.392}    & 0.455             & 0.451 & \underline{0.401} & \underline{0.410} & 0.483             & 0.479             & 0.402             & 0.414             & 0.477             & 0.476             & 0.430          & 0.439          & 0.512             & 0.493             & 0.429             & 0.439 & 0.520          & 0.504             \\
    \multicolumn{1}{c|}{}                                                           & 336   & \textbf{0.399}    & \textbf{0.428}    & 0.477             & 0.479 & \underline{0.437} & \underline{0.442} & 0.498             & 0.510             & 0.452             & 0.452             & 0.594             & 0.541             & 0.485          & 0.479          & 0.552             & 0.551             & 0.496             & 0.487 & 0.626          & 0.559             \\
    \multicolumn{1}{c|}{}                                                           & 720   & \textbf{0.426}    & \underline{0.457} & 0.467             & 0.490 & \underline{0.442} & \textbf{0.454}    & 0.556             & 0.527             & 0.462             & 0.468             & 0.831             & 0.657             & 0.500          & 0.497          & 0.562             & 0.560             & 0.463             & 0.474 & 0.863          & 0.672             \\
    \midrule                                                                                                                                                                                                                                                                                                                                                                                                                                                                                              
    \multicolumn{1}{c|}{\multirow{4}{*}{\begin{sideways}ECL\end{sideways}}}         & 96    & \textbf{0.152}    & \textbf{0.254}    & 0.182             & 0.297 & 0.211             & 0.312             & 0.187             & 0.283             & \underline{0.168} & \underline{0.272} & 0.197             & 0.282             & 0.187          & 0.304          & 0.169             & 0.273             & 0.193             & 0.308 & 0.207          & 0.307             \\
    \multicolumn{1}{c|}{}                                                           & 192   & \textbf{0.165}    & \textbf{0.263}    & 0.188             & 0.300 & 0.214             & 0.313             & 0.258             & 0.330             & 0.184             & 0.289             & 0.196             & \underline{0.285} & 0.199          & 0.315          & \underline{0.182} & 0.286             & 0.201             & 0.315 & 0.213          & 0.316             \\
    \multicolumn{1}{c|}{}                                                           & 336   & \textbf{0.173}    & \textbf{0.273}    & 0.210             & 0.324 & 0.230             & 0.328             & 0.323             & 0.369             & \underline{0.198} & \underline{0.300} & 0.209             & 0.301             & 0.212          & 0.329          & 0.200             & 0.304             & 0.214             & 0.329 & 0.230          & 0.333             \\
    \multicolumn{1}{c|}{}                                                           & 720   & \textbf{0.201}    & \textbf{0.299}    & 0.232             & 0.339 & 0.272             & 0.359             & 0.404             & 0.423             & \underline{0.220} & \underline{0.320} & 0.245             & 0.333             & 0.233          & 0.345          & 0.222             & 0.321             & 0.246             & 0.355 & 0.265          & 0.360             \\
    \midrule                                                                                                                                                                                                                                                                                                                                                                                                                                                                                              
    \multicolumn{1}{c|}{\multirow{4}{*}{\begin{sideways}Traffic\end{sideways}}}     & 96    & \textbf{0.500}    & 0.324             & 0.564             & 0.351 & 0.579             & 0.388             & \underline{0.512} & \textbf{0.290}    & 0.593             & \underline{0.321} & 0.650             & 0.396             & 0.607          & 0.392          & 0.612             & 0.338             & 0.587             & 0.366 & 0.615          & 0.391             \\
    \multicolumn{1}{c|}{}                                                           & 192   & \textbf{0.506}    & \underline{0.324} & 0.570             & 0.349 & 0.571             & 0.382             & \underline{0.523} & \underline{0.297} & 0.617             & 0.336             & 0.598             & 0.370             & 0.621          & 0.399          & 0.613             & 0.340             & 0.604             & 0.373 & 0.601          & 0.382             \\
    \multicolumn{1}{c|}{}                                                           & 336   & \textbf{0.528}    & 0.341             & 0.576             & 0.349 & 0.582             & 0.385             & \underline{0.530} & \textbf{0.300}    & 0.629             & \underline{0.336} & 0.605             & 0.373             & 0.622          & 0.396          & 0.618             & 0.328             & 0.621             & 0.383 & 0.613          & 0.386             \\
    \multicolumn{1}{c|}{}                                                           & 720   & \textbf{0.561}    & 0.369             & 0.602             & 0.360 & 0.596             & 0.389             & \underline{0.573} & \textbf{0.313}    & 0.640             & \underline{0.350} & 0.645             & 0.394             & 0.632          & 0.396          & 0.653             & 0.355             & 0.626             & 0.382 & 0.658          & 0.407             \\
    \midrule                                                                                                                                                                                                                                                                                                                                                                                                                                                                                              
    \multicolumn{1}{c|}{\multirow{4}{*}{\begin{sideways}Weather\end{sideways}}}     & 96    & \textbf{0.148}    & \textbf{0.212}    & 0.220             & 0.289 & 0.180             & 0.222             & \underline{0.153} & \underline{0.217} & 0.172             & 0.220             & 0.196             & 0.255             & 0.197          & 0.281          & 0.173             & 0.223             & 0.217             & 0.296 & 0.182          & 0.242             \\
    \multicolumn{1}{c|}{}                                                           & 192   & \underline{0.200} & 0.262             & 0.341             & 0.385 & 0.229             & \textbf{0.261}    & \textbf{0.197}    & 0.269             & 0.219             & \textbf{0.261}    & 0.237             & 0.296             & 0.237          & 0.312          & 0.245             & 0.285             & 0.276             & 0.336 & 0.227          & 0.287             \\
    \multicolumn{1}{c|}{}                                                           & 336   & \underline{0.256} & 0.310             & 0.463             & 0.455 & 0.281             & \textbf{0.298}    & \textbf{0.252}    & 0.311             & \underline{0.280} & \underline{0.306} & 0.283             & 0.335             & 0.298          & 0.353          & 0.321             & 0.338             & 0.339             & 0.380 & 0.282          & 0.334             \\
    \multicolumn{1}{c|}{}                                                           & 720   & \underline{0.327} & 0.362             & 0.682             & 0.565 & 0.358             & \underline{0.349} & \textbf{0.318}    & 0.363             & 0.365             & 0.359             & 0.345             & 0.381             & 0.352          & \textbf{0.288} & 0.414             & 0.410             & 0.403             & 0.428 & 0.352          & 0.386             \\
    \midrule                                                                                                                                                                                                                                                                                                                                                                                                                                                                                     
    \multicolumn{1}{c|}{\multirow{4}{*}{\begin{sideways}Exchange\end{sideways}}}    & 96    & \textbf{0.085}    & \underline{0.203} & 0.109             & 0.240 & \textbf{0.085}    & \textbf{0.202}    & 0.186             & 0.346             & 0.107             & 0.234             & 0.088             & 0.218             & \textbf{0.085} & 0.204          & 0.111             & 0.237             & 0.148             & 0.278 & 0.116          & 0.262             \\
    \multicolumn{1}{c|}{}                                                           & 192   & \textbf{0.176}    & \textbf{0.297}    & 0.241             & 0.353 & 0.180             & \underline{0.301} & 0.467             & 0.522             & 0.226             & 0.344             & \textbf{0.176}    & 0.315             & 0.182          & 0.303          & 0.219             & 0.335             & 0.271             & 0.380 & 0.215          & 0.359             \\
    \multicolumn{1}{c|}{}                                                           & 336   & \underline{0.336} & \textbf{0.418}    & 0.471             & 0.508 & \underline{0.336} & \underline{0.420} & 0.783             & 0.721             & 0.367             & 0.448             & \textbf{0.313}    & 0.427             & 0.348          & 0.428          & 0.421             & 0.476             & 0.460             & 0.500 & 0.377          & 0.466             \\
    \multicolumn{1}{c|}{}                                                           & 720   & 0.953             & 0.738             & 1.259             & 0.865 & 0.881             & 0.710             & 1.367             & 0.943             & 0.964             & 0.746             & \underline{0.839} & \textbf{0.695}    & 1.025          & 0.774          & 1.092             & 0.769             & 1.195             & 0.841 & \textbf{0.831} & \underline{0.699} \\
    \bottomrule
    \end{tabular}%
  \label{tab:ltf}%
\end{table*}%
\subsection{Long-Term Forecasting}
\label{sec:ltf}

\subsubsection{Task Settings}
Long-term time series forecasting has always been a primary goal of time series analysis. Characterized by its extraordinary long horizon as output, the forecasting testifies the long-range modeling ability of different schemes. In this task, we evaluate the approaches on eight popular real-world datasets across energy, transportation, weather, and finance domains. Each dataset contains one long multivariate time series. Information of the datasets is summarized in Table \ref{tab:ltfdata}.

We include Scaleformer \cite{shabani2023scaleformer} as a task-specific baseline in this experiment. Scaleformer is the latest Transformer-based approach with impressive performance in long-term forecasting by iteratively refining the forecasting result at multiple time scales.

In this task, we denote the model input as $\boldsymbol{X}\in\mathbb{R}^{C\times L}$, and the model output as $\boldsymbol{Y}\in\mathbb{R}^{C\times H}$, where $C$, $L$, and $H$ are the number of channels, the length of input time series, and the forecasting horizon, respectively. We fix the length of input time series as 96, and then train and test each scheme with four forecasting horizons, i.e., 96, 192, 336, and 720, to evaluate the performance of different schemes under different forecasting horizons. We obtain input and output sample pairs with a sliding window over the long time series. We use the mean squared error (MSE) between the ground truth $\boldsymbol{Y}$ and the prediction $\boldsymbol{\hat{Y}}$ as the loss function to train the models, and report both MSE and mean absolute error (MAE) for performance evaluation.

\subsubsection{Result Analysis}
\hl{As shown in Table~\ref{tab:ltf}, \shortname{} achieves the best performance on most datasets, including different forecasting horizon settings, with 45 first and 9 second places out of 64 benchmarks in total. Furthermore, on most benchmarks, \shortname{} outperforms the second place by a significant margin. From the results, we can see that although ETSformer adopts the decomposition design, it is still based on the pair-wise self-attention for temporal feature extraction, which has been shown to be ineffective for sub-series modeling. Therefore it struggles to perform well. TimesNet and PatchTST consider sub-series modeling in their design. Thus these two schemes are shown to be the strongest baselines. Compared with them, \shortname{} still outperforms significantly by combining decomposition with multi-scale sub-series modeling in the design. We believe that the outstanding performance of \shortname{} fully demonstrates the efficacy of our multi-scale decomposition in time series modeling.}

\subsection{Short-Term Forecasting}
\label{sec:stf}
\begin{table}
  \small
  \setlength\tabcolsep{3pt}
  \centering
  \caption{Statistics of datasets for short-term forecasting.}
    \begin{tabular}{ccccc}
    \toprule
    Dataset & Dim   & Length & Train & Test \\
    \midrule
    Yearly & 1     & 6     & 23000 & 23000 \\
    Quarterly & 1     & 8     & 24000 & 24000 \\
    Monthly & 1     & 18    & 48000 & 48000 \\
    Weekly & 1     & 13    & 359   & 359 \\
    Daily & 1     & 14    & 4227  & 4227 \\
    Hourly & 1     & 48    & 414   & 414 \\
    \bottomrule
    \end{tabular}%
  \label{tab:stfdata}%
\end{table}%

\begin{table*}[]
  \small
  \centering
  \renewcommand\arraystretch{0.75}
  \setlength\tabcolsep{3pt}
  \caption{\hl{Short-term forecasting results. The best results are in \textbf{bold} and the second bests are \underline{underlined}. (*Task-specific baselines.)}}
    \begin{tabular}{c|c|c|c|c|c|c|c|c|c|c|c|c}
    \toprule
    \multicolumn{2}{c|}{\multirow{2}{*}{Models}} & MSD-Mixer & N-HiTS* & N-BEATS* & PatchTST & Crossformer & TimesNet & DLinear & ETSformer & NST    & FEDformer & LightTS \\
    \multicolumn{2}{c|}{}                        & (Ours)    & (2023)  & (2020)   & (2023)   & (2023)      & (2023)   & (2023)  & (2022)    & (2022) & (2022)    & (2022)    \\
    \midrule                                                                                                                                                                           
    \multirow{3}{*}{\begin{sideways}Yr.\end{sideways}}   & SMAPE & \textbf{13.191} & 13.418            & 13.436            & 13.777 & 13.392 & \underline{13.387} & 16.965 & 18.009 & 13.717 & 13.728 & 14.247 \\
                                                         & MASE  & \textbf{2.967}  & 3.045             & 3.043             & 3.056  & 3.001  & \underline{2.996}  & 4.283  & 4.487  & 3.078  & 3.048  & 3.109  \\
                                                         & OWA   & \textbf{0.777}  & 0.793             & 0.794             & 0.806  & 0.787  & \underline{0.786}  & 1.058  & 1.115  & 0.807  & 0.803  & 0.827  \\
    \midrule                                                                                                                                                                                                        
    \multirow{3}{*}{\begin{sideways}Qtr.\end{sideways}}  & SMAPE & \textbf{9.971}  & 10.202            & 10.124            & 11.058 & 16.317 & \underline{10.100} & 12.145 & 13.376 & 10.958 & 10.792 & 11.364 \\
                                                         & MASE  & \textbf{1.151}  & 1.194             & \underline{1.169} & 1.321  & 2.197  & 1.182              & 1.520  & 1.906  & 1.325  & 1.283  & 1.328  \\
                                                         & OWA   & \textbf{0.872}  & 0.899             & \underline{0.886} & 0.984  & 1.542  & 0.890              & 1.106  & 1.302  & 0.981  & 0.958  & 1.000  \\
    \midrule                                                                                                                                                                                                      
    \multirow{3}{*}{\begin{sideways}Mon.\end{sideways}}  & SMAPE & \textbf{12.588} & 12.791            & 12.677            & 14.433 & 12.924 & \underline{12.670} & 13.514 & 14.588 & 13.917 & 14.260 & 14.014 \\
                                                         & MASE  & \textbf{0.921}  & 0.969             & 0.937             & 1.154  & 0.966  & \underline{0.933}  & 1.037  & 1.368  & 1.097  & 1.102  & 1.053  \\
                                                         & OWA   & \textbf{0.869}  & 0.899             & 0.880             & 1.043  & 0.902  & \underline{0.878}  & 0.956  & 1.149  & 0.998  & 1.012  & 0.981  \\
    \midrule                                                                                                                                                                                                      
    \multirow{3}{*}{\begin{sideways}Oth.\end{sideways}}  & SMAPE & \textbf{4.615}  & 5.061             & 4.925             & 5.216  & 5.493  & \underline{4.891}  & 6.709  & 7.267  & 6.302  & 4.954  & 15.880 \\
                                                         & MASE  & \textbf{3.124}  & \underline{3.216} & 3.391             & 3.688  & 3.690  & 3.302              & 4.953  & 5.240  & 4.064  & 3.264  & 11.434 \\
                                                         & OWA   & \textbf{0.978}  & 1.040             & 1.053             & 1.130  & 1.160  & \underline{1.035}  & 1.487  & 1.591  & 1.304  & 1.036  & 3.474  \\
    \midrule                                                                                                                                                                                                      
    \multirow{3}{*}{\begin{sideways}Avg.\end{sideways}}  & SMAPE & \textbf{11.700} & 11.927            & 11.851            & 13.011 & 13.474 & \underline{11.829} & 13.639 & 14.718 & 12.780 & 12.840 & 13.525 \\
                                                         & MASE  & \textbf{1.557}  & 1.613             & 1.599             & 1.758  & 1.866  & \underline{1.585}  & 2.095  & 2.408  & 1.756  & 1.701  & 2.111  \\
                                                         & OWA   & \textbf{0.838}  & 0.861             & 0.855             & 0.939  & 0.985  & \underline{0.851}  & 1.051  & 1.172  & 0.930  & 0.918  & 1.051  \\
    \bottomrule                                                                                                                                                                                    
    \end{tabular}%
  \label{tab:stf}%
\end{table*}%
\subsubsection{Task Settings}
In this task, we adopt the dataset and performance measures from the well-known M4 competition~\cite{makridakis2018m4}, which focuses on the short-term forecasting of univariate time series. The dataset contains 100,000 sequences of data in total, which are further divided into 6 subsets by sampling intervals including yearly, quarterly, monthly, weekly, daily, and hourly. More information of the dataset is summarized in Table~\ref{tab:stfdata}. Each subset contains real-life time series data from different domains, such as economics, finance, industry, demographics, etc. The analysis requires the forecasting models to learn the general temporal patterns from samples across diverse domains.

We include N-BEATS~\cite{oreshkin2020n} and N-HiTS~\cite{challu2023nhits} as task-specific baselines. N-BEATS is an MLP-based model for time series decomposition. N-HiTS further enhances N-BEATS with multi-scale modeling by introducing down-sampling and interpolation. They are proposed for univariate time series forecasting only. 

We evaluate with the symmetric mean absolute percentage error (SMAPE) and the mean absolute scaled error (MASE), defined as
\begin{equation}
  \begin{aligned}
    \mathrm{SMAPE}=&\frac{200}{H}\sum_{i=1}^{H}\frac{|y_i-\hat{y_i}|}{|y_i|+|\hat{y_i}|},\\
    \mathrm{MASE}=&\frac{1}{H}\sum_{i=1}^{H}\frac{|y_i-\hat{y_i}|}{\frac{1}{H-m}\sum_{j=m+1}^{H}|y_j-y_{h-m}|},
  \end{aligned}
\end{equation}
where $y_i$ and $\hat{y_i}$ are the ground truth and forecasting of the $i$-th time step in $H$ total future time steps, $m$ is the periodicity of the data. Intuitively, SMAPE measures the relative errors of the forecasting result. MASE is the MAE of the forecast values divided by the MAE of the in-sample one-step naive forecast. We also use the overall weighted average (OWA) of SMAPE and MASE as an evaluation metric, which is defined by the M4 competition~\cite{makridakis2018m4}.


\subsubsection{Result Analysis}
As shown in Table \ref{tab:stf}, \shortname{} again leads the board with top-1 performance in every benchmark, which demonstrates the excellent capability of \shortname{} on modeling short and univariate time series. Each time point in a univariate time series is a single scalar, which makes it even harder for Transformer based models to attain meaningful attention scores. Therefore, Transformer based approaches, which rely on pair-wise correlations have inferior performance in this task, especially ETSformer. Among the baselines, N-HiTS and N-BEATS are based on decomposition, and they show satisfactory performance in this task, which validates the effectiveness of decomposition for time series modeling. \shortname{} further advances N-BEATS and N-HiTS with \emph{multi-scale temporal patching} and mixing, as well as the \emph{residual loss} for better extraction of temporal patterns, which facilitates MSD-Mixer with stronger modeling ability than them, and helps \shortname{} achieve the best performance.

\subsection{Imputation}
\label{sec:imp}

\begin{table*}[]
  \small
  \centering
  \setlength\tabcolsep{3pt}
  \renewcommand\arraystretch{0.8}
  \caption{\hl{Imputation results. The best results are in \textbf{bold} and the second bests are \underline{underlined}.}}
    \begin{tabular}{cc|cc|cc|cc|cc|cc|cc|cc|cc|cc}
    \toprule
    \multicolumn{2}{c|}{\multirow{2}{*}{Models}} & \multicolumn{2}{c|}{MSD-Mixer} & \multicolumn{2}{c|}{PatchTST} & \multicolumn{2}{c|}{Crossformer} & \multicolumn{2}{c|}{TimesNet} & \multicolumn{2}{c|}{DLinear} & \multicolumn{2}{c|}{ETSformer} & \multicolumn{2}{c|}{NST}    & \multicolumn{2}{c|}{FEDformer} & \multicolumn{2}{c}{LightTS} \\
    \multicolumn{2}{c|}{}                        & \multicolumn{2}{c|}{(Ours)}    & \multicolumn{2}{c|}{(2023)}   & \multicolumn{2}{c|}{(2023)}      & \multicolumn{2}{c|}{(2023)}   & \multicolumn{2}{c|}{(2023)}  & \multicolumn{2}{c|}{(2022)}    & \multicolumn{2}{c|}{(2022)} & \multicolumn{2}{c|}{(2022)}    & \multicolumn{2}{c}{(2022)}  \\
    \midrule                                                                                                                                                                                                                                                                                                                                                                
    \multicolumn{2}{c|}{Metric} & MSE   & MAE   & MSE   & MAE   & MSE   & MAE   & MSE   & MAE   & MSE   & MAE   & MSE   & MAE   & MSE   & MAE & MSE   & MAE  & MSE   & MAE \\
    \midrule                                                                                                                                                                                                                                                                                                                                                                
    \multicolumn{1}{c|}{\multirow{4}{*}{\begin{sideways}ETTm1\end{sideways}}}   & 12.5\% & \textbf{0.019} & \underline{0.096} & 0.047 & 0.138 & 0.037             & 0.137             & \textbf{0.019}    & \textbf{0.092}    & 0.058 & 0.162 & 0.067 & 0.188 & 0.026             & 0.107             & 0.035 & 0.135 & 0.075 & 0.180 \\
    \multicolumn{1}{c|}{}                                                       & 25\%   & \textbf{0.019} & \textbf{0.092}    & 0.040 & 0.127 & 0.038             & 0.141             & \underline{0.023} & \underline{0.101} & 0.080 & 0.193 & 0.096 & 0.229 & 0.032             & 0.119             & 0.052 & 0.166 & 0.093 & 0.206 \\
    \multicolumn{1}{c|}{}                                                       & 37.5\% & \textbf{0.024} & \textbf{0.103}    & 0.043 & 0.132 & 0.041             & 0.142             & \underline{0.029} & 0.111             & 0.103 & 0.219 & 0.133 & 0.271 & 0.039             & \underline{0.131} & 0.069 & 0.191 & 0.113 & 0.231 \\
    \multicolumn{1}{c|}{}                                                       & 50\%   & \textbf{0.027} & \textbf{0.103}    & 0.048 & 0.139 & 0.047             & 0.152             & \underline{0.036} & \underline{0.124} & 0.132 & 0.248 & 0.186 & 0.323 & 0.047             & 0.145             & 0.089 & 0.218 & 0.134 & 0.255 \\
    \midrule                                                                                                                                                                                                                                                                                                                                                                                                                                   
    \multicolumn{1}{c|}{\multirow{4}{*}{\begin{sideways}ETTm2\end{sideways}}}   & 12.5\% & \textbf{0.018} & \textbf{0.079}    & 0.026 & 0.093 & 0.044             & 0.148             & \textbf{0.018}    & \underline{0.080} & 0.062 & 0.166 & 0.108 & 0.239 & 0.021             & 0.088             & 0.056 & 0.159 & 0.034 & 0.127 \\
    \multicolumn{1}{c|}{}                                                       & 25\%   & \textbf{0.020} & \textbf{0.084}    & 0.026 & 0.094 & 0.047             & 0.151             & \textbf{0.020}    & \underline{0.085} & 0.085 & 0.196 & 0.164 & 0.294 & 0.024             & 0.096             & 0.080 & 0.195 & 0.042 & 0.143 \\
    \multicolumn{1}{c|}{}                                                       & 37.5\% & \textbf{0.022} & \textbf{0.091}    & 0.033 & 0.110 & 0.044             & 0.145             & \underline{0.023} & \textbf{0.091}    & 0.106 & 0.222 & 0.237 & 0.356 & 0.027             & 0.103             & 0.110 & 0.231 & 0.051 & 0.159 \\
    \multicolumn{1}{c|}{}                                                       & 50\%   & \textbf{0.026} & \underline{0.100} & 0.033 & 0.106 & 0.047             & 0.150             & \textbf{0.026}    & \textbf{0.098}    & 0.131 & 0.247 & 0.323 & 0.421 & 0.030             & 0.108             & 0.156 & 0.276 & 0.059 & 0.174 \\
    \midrule                                                                                                                                                                                                                                                                                                                                                                                                                                
    \multicolumn{1}{c|}{\multirow{4}{*}{\begin{sideways}ETTh1\end{sideways}}}   & 12.5\% & \textbf{0.031} & \textbf{0.116}    & 0.081 & 0.189 & 0.099             & 0.218             & \underline{0.057} & \underline{0.159} & 0.151 & 0.267 & 0.126 & 0.263 & 0.060             & 0.165             & 0.070 & 0.190 & 0.240 & 0.345 \\
    \multicolumn{1}{c|}{}                                                       & 25\%   & \textbf{0.041} & \textbf{0.135}    & 0.093 & 0.202 & 0.125             & 0.243             & \underline{0.069} & \underline{0.178} & 0.180 & 0.292 & 0.169 & 0.304 & 0.080             & 0.189             & 0.106 & 0.236 & 0.265 & 0.364 \\
    \multicolumn{1}{c|}{}                                                       & 37.5\% & \textbf{0.056} & \textbf{0.157}    & 0.104 & 0.214 & 0.146             & 0.263             & \underline{0.084} & \underline{0.196} & 0.215 & 0.318 & 0.220 & 0.347 & 0.102             & 0.212             & 0.124 & 0.258 & 0.296 & 0.382 \\
    \multicolumn{1}{c|}{}                                                       & 50\%   & \textbf{0.071} & \textbf{0.179}    & 0.124 & 0.232 & 0.158             & 0.281             & \underline{0.102} & \underline{0.215} & 0.257 & 0.347 & 0.293 & 0.402 & 0.133             & 0.240             & 0.165 & 0.299 & 0.334 & 0.404 \\
    \midrule                                                                                                                                                                                                                                                                                                                                                                                                                                   
    \multicolumn{1}{c|}{\multirow{4}{*}{\begin{sideways}ETTh2\end{sideways}}}   & 12.5\% & \textbf{0.037} & \textbf{0.125}    & 0.059 & 0.152 & 0.103             & 0.220             & \underline{0.040} & \underline{0.130} & 0.100 & 0.216 & 0.187 & 0.319 & 0.042             & 0.133             & 0.095 & 0.212 & 0.101 & 0.231 \\
    \multicolumn{1}{c|}{}                                                       & 25\%   & \textbf{0.040} & \textbf{0.131}    & 0.059 & 0.154 & 0.110             & 0.229             & \underline{0.046} & \underline{0.141} & 0.127 & 0.247 & 0.279 & 0.390 & 0.049             & 0.147             & 0.137 & 0.258 & 0.115 & 0.246 \\
    \multicolumn{1}{c|}{}                                                       & 37.5\% & \textbf{0.048} & \textbf{0.145}    & 0.064 & 0.161 & 0.129             & 0.246             & \underline{0.052} & \underline{0.151} & 0.158 & 0.276 & 0.400 & 0.465 & 0.056             & 0.158             & 0.187 & 0.304 & 0.126 & 0.257 \\
    \multicolumn{1}{c|}{}                                                       & 50\%   & \textbf{0.058} & \underline{0.163} & 0.070 & 0.170 & 0.148             & 0.265             & \underline{0.060} & \textbf{0.162}    & 0.183 & 0.299 & 0.602 & 0.572 & 0.065             & 0.170             & 0.232 & 0.341 & 0.136 & 0.268 \\
    \midrule                                                                                                                                                                                                                                                                                                                                                                                                       
    \multicolumn{1}{c|}{\multirow{4}{*}{\begin{sideways}ECL\end{sideways}}}     & 12.5\% & \textbf{0.048} & \textbf{0.150}    & 0.103 & 0.215 & \underline{0.068} & \underline{0.181} & 0.085             & 0.202             & 0.092 & 0.214 & 0.196 & 0.321 & 0.093             & 0.210             & 0.107 & 0.237 & 0.102 & 0.229 \\
    \multicolumn{1}{c|}{}                                                       & 25\%   & \textbf{0.059} & \textbf{0.170}    & 0.105 & 0.219 & \underline{0.079} & \underline{0.198} & 0.089             & 0.206             & 0.118 & 0.247 & 0.207 & 0.332 & 0.097             & 0.214             & 0.120 & 0.251 & 0.121 & 0.252 \\
    \multicolumn{1}{c|}{}                                                       & 37.5\% & \textbf{0.070} & \textbf{0.184}    & 0.109 & 0.225 & \underline{0.087} & \underline{0.203} & 0.094             & 0.213             & 0.144 & 0.276 & 0.219 & 0.344 & 0.102             & 0.220             & 0.136 & 0.266 & 0.141 & 0.273 \\
    \multicolumn{1}{c|}{}                                                       & 50\%   & \textbf{0.080} & \textbf{0.197}    & 0.113 & 0.231 & \underline{0.113} & \underline{0.212} & 0.100             & 0.221             & 0.175 & 0.305 & 0.235 & 0.357 & 0.108             & 0.228             & 0.158 & 0.284 & 0.160 & 0.293 \\
    \midrule                                                                                                                                                                                                                                                                                                                                                                                                           
    \multicolumn{1}{c|}{\multirow{4}{*}{\begin{sideways}Weather\end{sideways}}} & 12.5\% & \textbf{0.025} & \textbf{0.043}    & 0.043 & 0.069 & 0.036             & 0.092             & \textbf{0.025}    & \underline{0.045} & 0.039 & 0.084 & 0.057 & 0.141 & 0.027             & 0.051             & 0.041 & 0.107 & 0.047 & 0.101 \\
    \multicolumn{1}{c|}{}                                                       & 25\%   & \textbf{0.028} & \textbf{0.050}    & 0.041 & 0.065 & 0.035             & 0.088             & \underline{0.029} & 0.052             & 0.048 & 0.103 & 0.065 & 0.155 & \underline{0.029} & 0.056             & 0.064 & 0.163 & 0.052 & 0.111 \\
    \multicolumn{1}{c|}{}                                                       & 37.5\% & \textbf{0.030} & \textbf{0.049}    & 0.043 & 0.069 & 0.035             & 0.088             & \underline{0.031} & \underline{0.057} & 0.057 & 0.117 & 0.081 & 0.180 & 0.033             & 0.062             & 0.107 & 0.229 & 0.058 & 0.121 \\
    \multicolumn{1}{c|}{}                                                       & 50\%   & \textbf{0.033} & \textbf{0.056}    & 0.045 & 0.070 & 0.038             & 0.092             & \underline{0.034} & \underline{0.062} & 0.066 & 0.134 & 0.102 & 0.207 & 0.037             & 0.068             & 0.183 & 0.312 & 0.065 & 0.133 \\
    \bottomrule
    \end{tabular}%
  \label{tab:imp}%
\end{table*}%

\subsubsection{Task Settings}
Missing values are common in real-world continuous data systems that collect time series data from various sources. A single missing data in a time series can break down the whole downstream application since most analysis methods assume complete data, which makes missing data imputation critical for time series analysis. In this task, we experiment on the ETT, ECL, and Weather datasets which are summarized in Table~\ref{tab:ltfdata}.

In the experiments, we first obtain the data samples $\boldsymbol{X}\in\mathbb{R}^{C\times L}$ with a sliding window, whose size $L$ is set as 96. We then create missing values $\boldsymbol{X}_{mask}$ in the data by randomly masking the $\boldsymbol{X}$ with zeros. We use $\boldsymbol{X}_{mask}$ as model input and the unmasked data $\boldsymbol{X}$ as ground truth for training and testing. To evaluate the performance under different ratios of missing data, for each dataset we train and evaluate each model with four missing data ratios, i.e., 12.5\%, 25\%, 37.5\%, and 50\%. We use the MSE between the ground truths and the predictions at the masked positions as the loss function to train the models, and report both MSE and MAE for performance evaluation. In particular, due to the missing data in the input, it is not feasible to compute the autocorrelation of the residual for \shortname{}. Therefore, we only compute the first term of the residual in the \emph{residual loss} (Equation~\ref{eq:acf_loss}).

\subsubsection{Result Analysis}
Results in Table \ref{tab:imp} show that \shortname{} also achieves the best performance on most datasets, as well as on different missing data ratios, with 45 first place out of 48 benchmarks in total. This task requires the model to learn correct temporal patterns from the data with missing values masked as zeros, which is challenging for most models. We observe that \shortname{} and TimesNet, which shows good performance in this task, have both considered sub-series modeling in multiple timescales. From this, we think sub-series modeling may help provide local context information for the estimation of missing values. Meanwhile, MSD-Mixer performs much better than TimesNet. This is because MSD-Mixer also considers multi-scale decomposition of the time series. It disentangles the temporal patterns within the data, such that they can be better modeled for the estimation of the missing value. Furthermore, the performance of other baseline methods drops quickly as the missing ratio increases, whereas the performance of our MSD-Mixer remains more stable, and consistently better than others. This also highlights the excellent capability of MSD-Mixer to model temporal patterns in complex time series data.

\subsection{Anomaly Detection}
\label{sec:ad}
\begin{table}
  \centering
  \small
  \renewcommand\arraystretch{0.9}
  \caption{Statistics of datasets for anomaly detection.}
    \begin{tabular}{ccccc}
    \toprule
    Dataset & Dim   & Length & Train & Test \\
    \midrule
    SMD   & 38    & 100   & 708405 & 708420 \\
    MSL   & 55    & 100   & 58317 & 73729 \\
    SMAP  & 25    & 100   & 135183 & 427617 \\
    SWaT  & 51    & 100   & 495000 & 449919 \\
    PSM   & 25    & 100   & 132481 & 87841 \\
    \bottomrule
    \end{tabular}%
  \label{tab:addata}%
\end{table}%

\begin{table*}[]
  \small
  \centering
  \setlength\tabcolsep{3pt}
  \renewcommand\arraystretch{0.8}
  \caption{\hl{Anomaly detection results. The best results are in \textbf{bold} and the second bests are \underline{underlined}. (*Task-specific baseline.)}}
    \begin{tabular}{c|c|c|c|c|c|c|c|c|c|c|c}
    \toprule
    \multicolumn{2}{c|}{\multirow{2}{*}{Models}} & MSD-Mixer & Anomaly* & PatchTST & Crossformer & TimesNet & DLinear & ETSformer & NST    & FEDformer & LightTS \\
    \multicolumn{2}{c|}{}                        & (Ours)    & (2022)   & (2023)   & (2023)      & (2023)   & (2023)  & (2022)    & (2022) & (2022)    & (2022)  \\
    \midrule                                                                                                                                                                                                
    \multirow{3}{*}{SMD}  & Precision & 88.7          & 88.9  & 87.5  & 83.1  & 88.7              & 83.6  & 87.4  & 88.3              & 88.0  & 87.1              \\
                          & Recall    & 86.1          & 82.2  & 82.2  & 76.6  & 83.1              & 71.5  & 79.2  & 81.2              & 82.4  & 78.4              \\
                          & F1-score  & \textbf{87.4} & 85.5  & 84.7  & 79.7  & \underline{85.8}  & 77.1  & 83.1  & 84.6              & 85.1  & 82.5              \\
    \midrule                                                                                                                                                                                                          
    \multirow{3}{*}{MSL}  & Precision & 91.3          & 79.6  & 87.4  & 84.7  & 83.9              & 84.3  & 85.1  & 68.6              & 77.1  & 82.4              \\
                          & Recall    & 88.4          & 87.4  & 69.5  & 83.7  & 86.4              & 85.4  & 84.9  & 89.1              & 80.1  & 75.8              \\
                          & F1-score  & \textbf{89.8} & 83.3  & 77.4  & 84.2  & \underline{85.2}  & 84.9  & 85.0  & 77.5              & 78.6  & 79.0              \\
    \midrule                                                                                                                                                                                                                                 
    \multirow{3}{*}{SMAP} & Precision & 93.4          & 91.9  & 90.5  & 92.0  & 92.5              & 92.3  & 92.3  & 89.4              & 90.5  & 92.6              \\
                          & Recall    & 96.9          & 58.1  & 56.4  & 55.4  & 58.3              & 55.4  & 55.8  & 59.0              & 58.1  & 55.3              \\
                          & F1-score  & \textbf{95.2} & 71.2  & 69.5  & 69.1  & \underline{71.5}  & 69.3  & 69.5  & 71.1              & 70.8  & 69.2              \\
    \midrule                                                                                                                                                                                                                                 
    \multirow{3}{*}{SWaT} & Precision & 93.1          & 72.5  & 91.3  & 88.5  & 86.8              & 80.9  & 90.0  & 68.0              & 90.2  & 92.0              \\
                          & Recall    & 98.3          & 97.3  & 83.2  & 93.5  & 97.3              & 95.3  & 80.4  & 96.8              & 96.4  & 94.7              \\
                          & F1-score  & \textbf{95.7} & 83.1  & 87.1  & 90.9  & 91.7              & 87.5  & 84.9  & 79.9              & 93.2  & \underline{93.3}  \\
    \midrule                                                                                                                                                                                                                     
    \multirow{3}{*}{PSM}  & Precision & 97.4          & 68.4  & 98.9  & 97.2  & 98.2              & 98.3  & 99.3  & 97.8              & 97.3  & 98.4              \\
                          & Recall    & 96.7          & 94.7  & 92.4  & 89.7  & 96.8              & 89.3  & 85.3  & 96.8              & 97.2  & 96.0              \\
                          & F1-score  & 97.0          & 79.4  & 95.6  & 93.3  & \textbf{97.5}     & 93.6  & 91.8  & \underline{97.3}  & 97.2  & 97.2              \\
    \bottomrule
    \end{tabular}%
  \label{tab:ad}%
\end{table*}%

\subsubsection{Task Settings}
Anomaly detection for time series data is of immense value in many real-time monitoring applications. It is also challenging due to the lack of labeled data. In this task, we leverage the popular paradigm of reconstruction-based unsupervised framework for anomaly detection. On this premise, a model learns to represent and reconstruct the normal data, thus abnormal data points can be detected with large reconstruction errors. Therefore, it is critical to learn high quality representations with the model. We experiment on five widely-adopted anomaly detection datasets for time series analysis, whose information is summarized in Table~\ref{tab:addata}.

We include the Anomaly Transformer~\cite{xu2022anomaly} as a task-specific baseline for comparison. Anomaly Transformer is one of the latest Transformer-based methods tailor-made for reconstruction-based unsupervised anomaly detection. It proposes a special Anomaly-Attention mechanism and a minimax strategy to learn and amplify the normal-abnormal associations.

For the experiment, we preprocess the datasets by splitting the time series into non-overlapping segments. In the training phase, we train the model to represent and reconstruct the input by minimizing reconstruction loss, which is the MSE between the model input and output. In the testing phase, we compute the difference between the test reconstruction loss of a data point and the average training reconstruction loss. If the difference is higher than a threshold, the data point is treated as an anomaly. The threshold values for different datasets are set as those in~\cite{wu2023timesnet}. We report the point-wise precision, recall, and F1-score of the detection results, and use F1-score to compare the performance of different methods.

\subsubsection{Result Analysis}
As shown in Table~\ref{tab:ad}, \shortname{} achieves the best F1-scores in 4 out of the 5 datasets. Compared with the baselines that simply learn to represent and reconstruct the time series, \shortname{} further learns to explicitly decompose the time series into components and represent each component for the reconstruction. Therefore, \shortname{} has a stronger representation learning ability to precisely capture the normal temporal patterns and identify the abnormal data.

\subsection{Classification}
\label{sec:cls}
\begin{table}
  \small
  \centering
  \renewcommand\arraystretch{0.8}
  \caption{Statistics of datasets for classification.}
    \begin{tabular}{cccccc}
    \toprule
    Dataset & Dim   & Length & Classes & Train & Test \\
    \midrule
    AWR   & 9     & 144   & 25    & 275   & 300 \\
    AF    & 2     & 640   & 3     & 15    & 15 \\
    CT    & 3     & 182   & 20    & 1,422 & 1,436 \\
    CR    & 6     & 1,197 & 12    & 108   & 72 \\
    FD    & 144   & 62    & 2     & 5,890 & 3,524 \\
    FM    & 28    & 50    & 2     & 316   & 100 \\
    MI    & 64    & 3,000 & 2     & 278   & 100 \\
    SCP1  & 6     & 896   & 2     & 268   & 293 \\
    SCP2  & 7     & 1,152 & 2     & 200   & 180 \\
    UWGL  & 3     & 315   & 8     & 120   & 320 \\
    \bottomrule
    \end{tabular}%
  \label{tab:clsdata}%
\end{table}%
\begin{table*}
  \centering
  \footnotesize
  \setlength\tabcolsep{1.5pt}
  \renewcommand\arraystretch{0.8}
  \caption{\hl{Classification results. The best results are in \textbf{bold} and the second bests are \underline{underlined}. (*Task-specific baselines.)}}
    \begin{tabular}{c|ccccccccccccccc}
    \toprule
    \multirow{2}{*}{Models} & MSD-Mixer & TARNet* & DTWD*  & TapNet* & MiniRocket* & TST*   & FormerTime* & PatchTST & Crossformer & TimesNet & DLinear & ETSformer & NST    & FEDformer & LightTS \\
                            & (Ours)    & (2022)  & (2015) & (2020)  & (2021)      & (2021) & (2023)      & (2023)   & (2023)      & (2023)   & (2023)  & (2022)    & (2022) & (2022)    & (2022)  \\
    \midrule
    AWR       & 0.983             & 0.977             & \underline{0.987} & \underline{0.987} & \textbf{0.993} & 0.947          & 0.985             & 0.040 & 0.937 & 0.977 & 0.963 & 0.973 & 0.497             & 0.587 & 0.970 \\
    AF        & \underline{0.600} & \textbf{1.000}    & 0.220             & 0.333             & 0.133          & 0.533          & \underline{0.600} & 0.467 & 0.400 & 0.333 & 0.200 & 0.400 & 0.467             & 0.400 & 0.333 \\
    CT        & 0.987             & \underline{0.994} & 0.989             & \textbf{0.997}    & 0.990          & 0.971          & 0.991             & 0.877 & 0.970 & 0.974 & 0.973 & 0.978 & 0.804             & 0.960 & 0.977 \\
    CR        & \textbf{1.000}    & \textbf{1.000}    & \textbf{1.000}    & 0.958             & 0.986          & 0.847          & 0.981             & 0.083 & 0.846 & 0.847 & 0.861 & 0.861 & 0.736             & 0.472 & 0.847 \\
    FD        & \textbf{0.698}    & 0.641             & 0.529             & 0.556             & 0.612          & 0.625          & \underline{0.687} & 0.500 & 0.687 & 0.686 & 0.672 & 0.673 & 0.500             & 0.684 & 0.658 \\
    FM        & \textbf{0.660}    & \underline{0.620} & 0.530             & 0.530             & 0.550          & 0.590          & 0.618             & 0.510 & 0.510 & 0.590 & 0.570 & 0.590 & 0.510             & 0.540 & 0.540 \\
    MI        & \textbf{0.670}    & 0.630             & 0.500             & 0.590             & 0.610          & 0.610          & 0.632             & 0.570 & 0.570 & 0.570 & 0.620 & 0.590 & \underline{0.640} & 0.580 & 0.590 \\
    SCP1      & \underline{0.949} & 0.816             & 0.775             & 0.652             & 0.915          & \textbf{0.961} & 0.887             & 0.741 & 0.921 & 0.918 & 0.880 & 0.860 & 0.898             & 0.594 & 0.918 \\
    SCP2      & \textbf{0.639}    & \underline{0.622} & 0.539             & 0.550             & 0.506          & 0.604          & 0.592             & 0.500 & 0.583 & 0.572 & 0.527 & 0.561 & 0.500             & 0.511 & 0.522 \\
    UWGL      & 0.884             & 0.878             & \underline{0.903} & 0.894             & 0.785          & \textbf{0.913} & 0.888             & 0.213 & 0.853 & 0.853 & 0.812 & 0.825 & 0.703             & 0.453 & 0.831 \\
    \midrule                                                                                                                                                                                                                          
    Avg. Acc. & \underline{0.807} & \textbf{0.818}    & 0.697             & 0.705             & 0.708          & 0.760          & 0.786             & 0.450 & 0.728 & 0.732 & 0.708 & 0.731 & 0.625             & 0.578 & 0.719 \\
    1st Count & \textbf{5}        & 3                 & 1                 & 1                 & 1              & 2              & 0                 & 0     & 0     & 0     & 0     & 0     & 0                 & 0     & 0     \\
    Mean Rank & \textbf{2.5}      & 4.4               & 8.3               & 7.5               & 8.0            & 6.0            & 3.8               & 13.0  & 8.5  & 7.1   & 9.0    & 7.4   & 11.0              & 11.3  & 8.5   \\
    \bottomrule
    \end{tabular}%
  \label{tab:cls}%
\end{table*}%

\begin{table*}
  \centering
  \small
  \renewcommand\arraystretch{0.76}
  \caption{Average results of \shortname{} variants on five tasks.}
  \begin{tabular}{cc|ccccc}
    \toprule
    \multicolumn{2}{c|}{Model} & MSD-Mixer & MSD-Mixer-I & MSD-Mixer-N & MSD-Mixer-U & MSD-Mixer-L \\
    \midrule
    \multirow{2}{*}{Long-Term Forecasting} & MSE   & 0.345 & 0.345 & 0.358 & 0.422 & 0.348 \\
          & MAE   & 0.358 & 0.357 & 0.371 & 0.470 & 0.360 \\
    \midrule
    \multirow{3}{*}{Short-Term Forecasting} & SMAPE & 11.700 & 11.699 & 11.814 & 11.869 & 11.780 \\
          & MASE  & 1.557 & 1.557 & 1.598 & 1.587 & 1.567 \\
          & OWA   & 0.838 & 0.837 & 0.853 & 0.853 & 0.844 \\
    \midrule
    \multirow{2}{*}{Imputation} & MSE   & 0.038 & 0.039 & 0.041 & 0.058 & 0.040 \\
          & MAE   & 0.117 & 0.130 & 0.122 & 0.149 & 0.119 \\
    \midrule
    Anomaly Detection & F1    & 0.930 & 0.925 & 0.918 & 0.847 & 0.897 \\
    \midrule
    Classification & ACC   & 0.807 & 0.803 & 0.732 & 0.729 & 0.768 \\
    \bottomrule
    \end{tabular}%
  \label{tab:abl}%
\end{table*}%

\subsubsection{Task Settings}
The time series classification problem arises in various real-life applications such as human activity recognition and medical time series based diagnosis. In this task we consider the series-level classification problem and build predictive models that output one categorical label for each time series, which emphasizes more on discriminative modeling ability than other time series analysis tasks. We experiment on ten datasets from the well-known UEA time series classification archive \cite{bagnall2018uea} which is the most widely used multivariate time series classification benchmark. The ten datasets have diverse characteristics in terms of domain, series length, number of samples, and the number of classes, which helps to comprehensively examine the capabilities of different methods. The datasets have been well processed and split into train and test sets. Information of the datasets is summarized in Table \ref{tab:clsdata}.

We include six competitive classification methods reported in recent works~\cite{cheng2023formertime,chowdhury2022tarnet} as task-specific baselines for comparison in this task. They cover both statistical (DTWD~\cite{shokoohi2015non}, MiniRocket~\cite{dempster2021minirocket}) and deep learning-based (TARNet~\cite{chowdhury2022tarnet}, FormerTime~\cite{cheng2023formertime}, TST~\cite{zerveas2021transformer}, TapNet~\cite{Zhang2020tapnet}) approaches. We use the classification accuracy, number of 1st counts and mean rank as our evaluation metrics. 

\subsubsection{Result Analysis}
Table~\ref{tab:cls} shows the result of our classification tests. \shortname{} performs the best with 5 first places and 2 second places out of 10 benchmarks, which demonstrates its great discriminative power in the modeling. Moreover, the diversity of datasets in terms of size, dimension, length, and number of classes also reflects the adaptability of \shortname{}. Different from other tasks discussed above, the task-general baselines typically perform inferior to task-specific ones in classification. The two best baselines are task-specific approaches TARNet and TST, which are Transformer-based deep learning algorithms. It should be noted that TARNet and TST adopt extra self-supervised training in addition to the supervised training with class labels, which we think may be the reason for their good performance. In contrast, \shortname{} outperforms them with only supervised training, which demonstrates the modeling ability of \shortname{}. We also notice that the two statistical baselines DTWD and MiniRocket perform well in some datasets. These results indicate that it is challenging to design a task-general backbone for classification tasks. We believe our thorough consideration on the special composition and multi-scale nature of time series is the reason why \shortname{} can consistently have better performance in classification.

\subsection{Ablation Study}
\label{sec:abl}
We strongly believe that the advantages of \shortname{} are rooted in our proposed \emph{multi-scale temporal patching} and \emph{residual loss}. To validate the efficacy of the proposed modules, we implement the following variants of \shortname{}:
\begin{itemize}
  \item \emph{\shortname{}-I}: the inverted \shortname{}. We arrange the layers with their patch sizes in ascending order instead of descending.
  \item \emph{\shortname{}-N}: the \shortname{} without patching. We replace the patching module with max pooling and linear interpolation layers, following the strategy in~\cite{challu2023nhits}.
  \item \emph{\shortname{}-U}: the \shortname{} without multi-scale patching. We set the patch size as the square root of the input length and use this same patch size for all layers.
  \item \emph{\shortname{}-L}: the \shortname{} trained without the \emph{residual loss}. We train the model with the loss function from the target task only.
\end{itemize}

\begin{figure*}
  \centering
  \includegraphics[width=\linewidth]{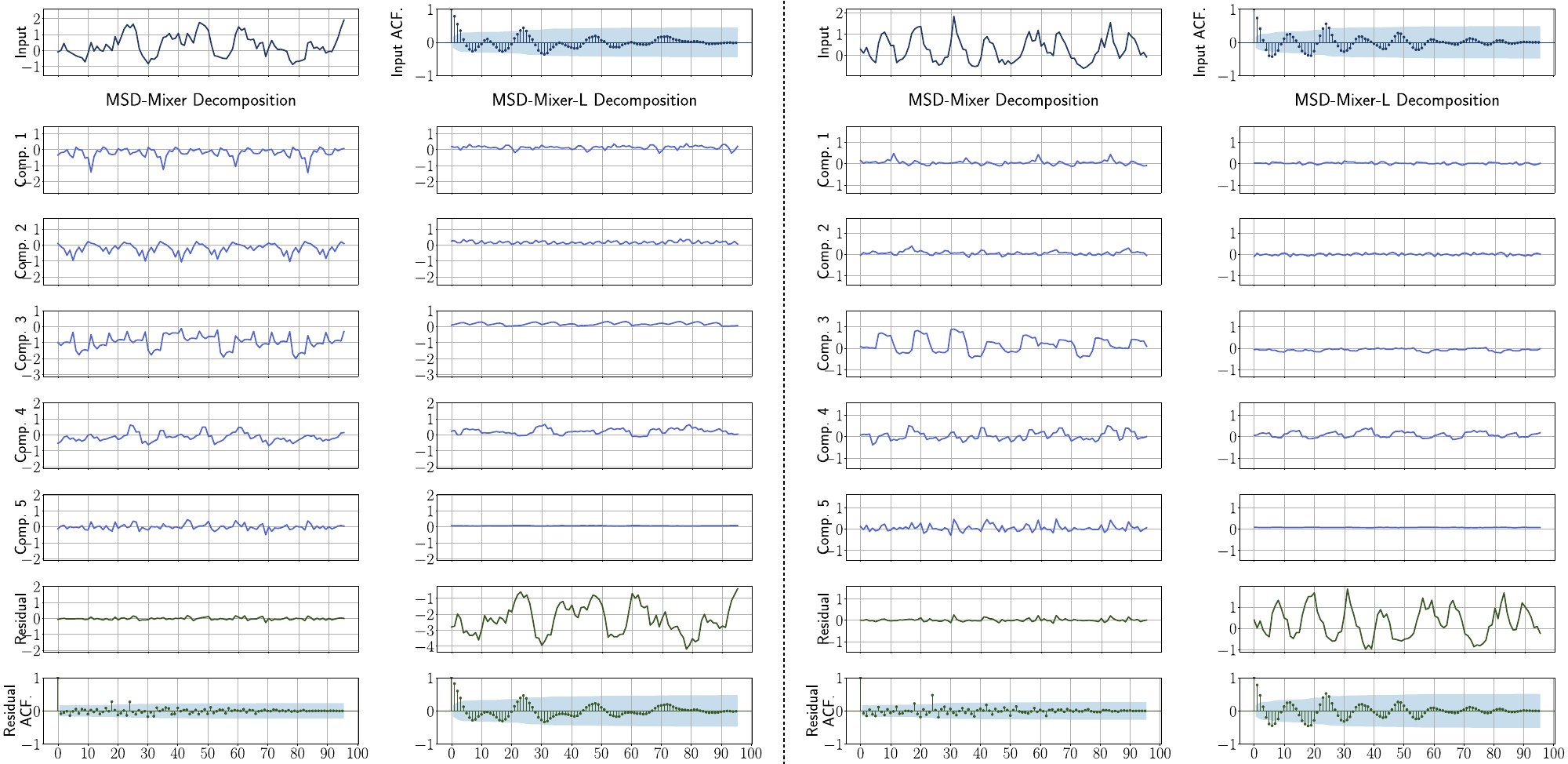}
  \caption{Examples of decomposition.}
  \label{fig:vis}
\end{figure*}

We carry out the experiments for the four \shortname{} variants on all benchmarks in the five tasks, and report their average performance over the benchmarks in each task in Table~\ref{tab:abl}. \shortname{}-I has very similar performance to the original \shortname{} for all five tasks. This result indicates that the arrangement of layers with different patch sizes does not affect the performance of \shortname{}. We think the reason behind is that the \emph{multi-scale temporal patching} enforces the layer to focus on the modeling of specific timescales, such that their order has a relatively small impact on the performance. Without the patching modules, \shortname{}-N cannot capture the sub-series features, thus we can observe a performance drop compared with \shortname{}, especially in classification accuracy. Likewise, by using the same patch size in all layers, \shortname{}-U does not model the multi-scale patterns in different layers, which affects the performance considerably in all tasks. Lastly, by comparing \shortname{}-L with \shortname{} we find that the \emph{residual loss} do contribute to the learning of the model in all tasks by enhancing the completeness of the decomposition.

\subsection{Model Efficiency}
\label{sec:eff}
\begin{figure}
  \centering
  \includegraphics[width=0.92\linewidth]{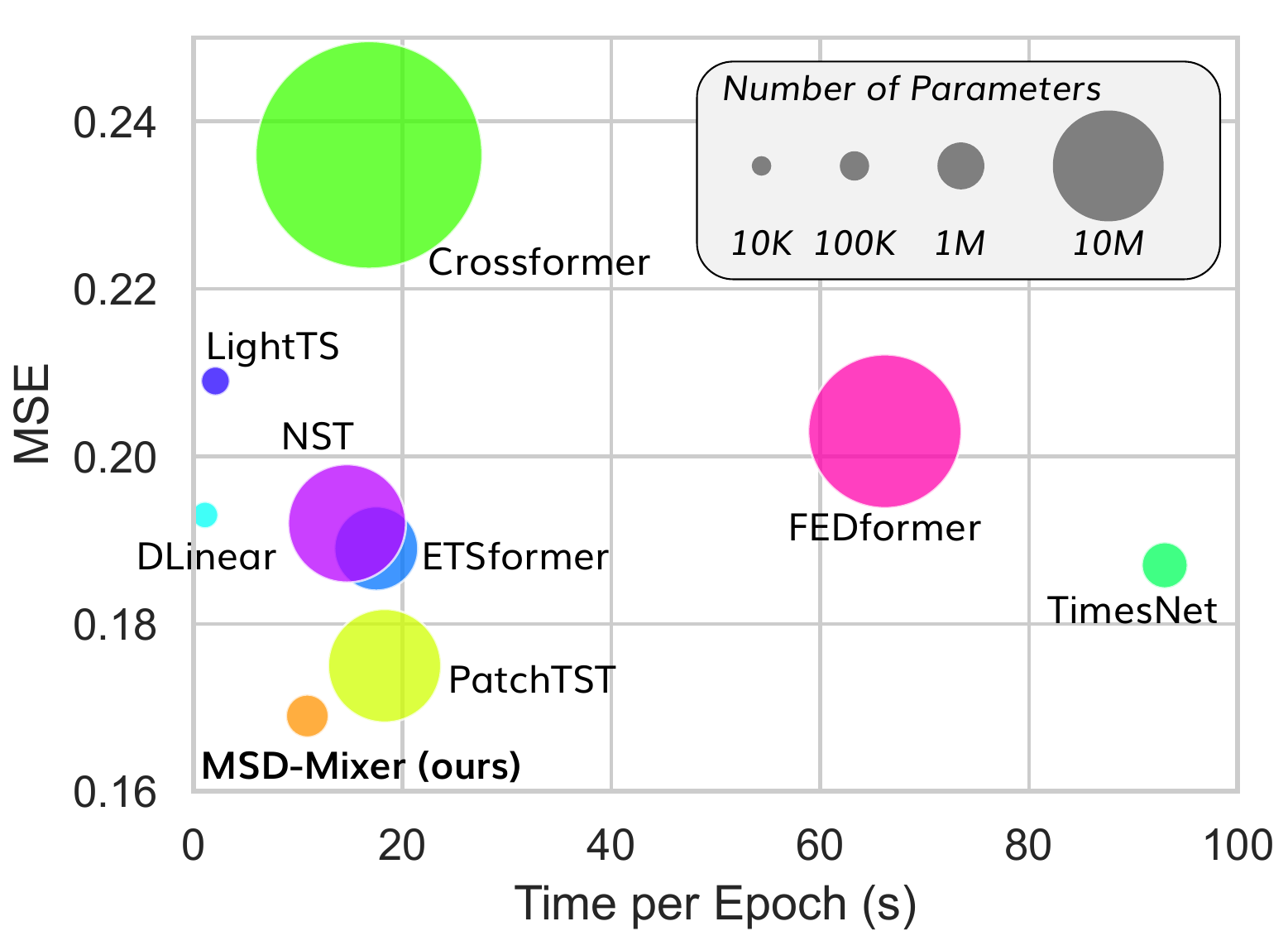}
  \caption{\hl{Model efficiency comparison.}}
  \label{fig:eff}
\end{figure}

\hl{To study the efficiency of our proposed \shortname{}, we compare the number of model parameters and training time consumption of different approaches in the long-term forecasting task with the ETTm2 dataset in Figure~\ref{fig:eff}. From the result we can observe that \shortname{} achieves the best MSE among all the baselines. Comparing with PatchTST and TimesNet which are the second- and third-best models, \shortname{} contains less than $1/10$ and $4/5$ model parameters (951K vs. 10.1M and 1191K), and runs more than $1.67$ and $8$ times faster (10.9s/epoch vs. 18.3s/epoch and 93.2s/epoch), which demonstrates the great efficiency of \shortname{}. On the other hand, MLP models (\shortname{}, DLinear, and LightTS) generally contain fewer model parameters and consume less training time than their Transformer and CNN counterparts in this experiment. Although \shortname{} is larger and slower than the other two MLP models, it achieves 12\% and 19\% improvements on MSE with the extra model parameters and time, which also shows \shortname{}'s advancements over the previous MLP-based methods.}
\subsection{Case Study}
\label{sec:cs}

To further validate the effectiveness of our carefully designed \emph{residual loss} in \shortname{}, in Figure \ref{fig:vis}, we show two examples of how the input time series is decomposed by \shortname{} trained with (\shortname{}) and without (\shortname{}-L) our proposed \emph{residual loss}. The examples are from the long-term forecasting task with the ETTh1 dataset, which is well acknowledged as a challenging dataset with complex characteristics including but not limited to multiple periodic variations and channel-wise heterogeneity. The sampling rate of the data is 1 hour, and input length is set to 96. We train the \shortname{} which has 5 layers with patch sizes as \{24, 12, 6, 2, 1\}, corresponding to sub-series of 1 day, half day, 6 hours, 2 hours, and 1 hour.

First, from both input plots we observe multiple irregular temporal patterns, which cannot be simply explained by seasonal or trend-cyclic patterns as discussed in previous works \cite{wu2021autoformer}. Their corresponding autocorrelation function (ACF) plots also indicate high correlations in multiple temporal lags in the input data. Therefore, simply considering seasonal-trend decomposition is not enough to account for intricate temporal patterns in real-life time series data. Then, we note that the components output by \shortname{} are obviously more diverse, especially in terms of their timescale, compared with that of \shortname{}-L. It indicates that they contain different temporal patterns in the input data. We attribute it to the effectiveness of our proposed \emph{multi-scale temporal patching}.

Furthermore, it is obvious from both examples that without our proposed \emph{residual loss}, \shortname{}-L leaves most of the information from the input in the decomposition residual, while the other components contain little information. In comparison, the mean of residuals from \shortname{} are much smaller, and the residual ACF plots also indicate less periodic patterns. The results clearly validate the effectiveness of our proposed \emph{residual loss} in constraining the decomposition residual. The multi-scale components also show the potential to provide interpretability on the composition of the input data and how the output is produced by \shortname{}.

\section{Conclusion}
\label{sec:conclu}
In this work, we solve the time series analysis problem by considering its unique composition and complex multi-scale temporal variations, and propose \shortname{}, a \longname{} which learns to explicitly decompose the input time series into different components, and represents the components in different layers. We propose a novel \emph{multi-scale temporal patching} approach in \shortname{} to model the time series as multi-scale patches, and employ MLPs along different dimensions to mix intra- and inter-patch variations and channel-wise correlations. In addition, we propose a \emph{residual loss} to constrain both the mean and the autocorrelation of the decomposition residual for decomposition completeness. Through extensive experiments on 26 real-world datasets, we demonstrate that \shortname{} consistently outperforms the state-of-the-art task-general and task-specific approaches by a wide margin on five common tasks, namely long-term forecasting (up to 9.8\% in MSE),  short-term forecasting (up to 5.6\% in OWA), imputation (up to 46.1\% in MSE), anomaly detection (up to 33.1\% in F1-score) and classification (up to 36.3\% in Mean Rank).

\section*{Acknowledgment}
The work of Weipeng Zhuo was supported in part by the Guangdong Provincial Key Laboratory IRADS
(2022B1212010006, R0400001-22).

\bibliographystyle{ACM-Reference-Format}
\bibliography{ref}
\end{document}